\documentclass[10pt,twocolumn,letterpaper]{article}

\usepackage{iccv}
\usepackage{times}
\usepackage{epsfig}
\usepackage{graphicx}
\usepackage{amsmath}
\usepackage{amssymb}
\usepackage{subfigure}
\usepackage{stfloats}
\usepackage{booktabs}
\usepackage{float}
\usepackage{graphicx}

\usepackage{bm}
\usepackage{caption}
\usepackage{color}
\usepackage{subfigure}
\usepackage{lipsum}
\usepackage{epstopdf}
\usepackage{array}
\usepackage{overpic}
\usepackage{mathtools}
\usepackage{rotating}
\usepackage{multirow}
\usepackage{bbding}
\usepackage{ulem}
\usepackage{ragged2e}

\usepackage{multirow}
\usepackage{makecell}
\usepackage{hhline}

\usepackage[american]{babel} 
\usepackage{cite}
\usepackage{cuted}
\usepackage[pagebackref=true,breaklinks=true,letterpaper=true,colorlinks,bookmarks=false]{hyperref}

\newcommand{\tabincell}[2]{\begin{tabular}{@{}#1@{}}#2\end{tabular}}

\iccvfinalcopy 

\def\iccvPaperID{5565} 

\ificcvfinal\fi

\begin{document}

\title{Learning RAW-to-sRGB Mappings with Inaccurately Aligned Supervision}

\author{{Zhilu Zhang$^{1}$, Haolin Wang$^{1}$, Ming Liu$^{1}$, Ruohao Wang$^{1}$, Jiawei Zhang$^{2}$, Wangmeng Zuo$^{1, 3}$ $^{(}$\Envelope$^)$ } \\
	$^1${Harbin Institute of Technology}, \ \ $^2${SenseTime Research}, \ \ $^3${Pazhou Lab, Guangzhou} \\
	{\tt\small{ \{cszlzhang, Why\_cs, csmliu, rhwangHIT\}@outlook.com} } {\tt {\small\{zhjw1988\}@gmail.com} } {\tt{\small{wmzuo@hit.edu.cn}}} 
}

\maketitle
\ificcvfinal\thispagestyle{empty}\fi

%
\begin{abstract}
%
  Learning RAW-to-sRGB mapping has drawn increasing attention in recent years, wherein an input raw image is trained to imitate the target sRGB image captured by another camera.
  However, the severe color inconsistency makes it very challenging to generate well-aligned training pairs of input raw and target sRGB images.
  While learning with inaccurately aligned supervision is prone to causing pixel shift and producing blurry results.
  In this paper, we circumvent such issue by presenting a joint learning model for image alignment and RAW-to-sRGB mapping.
  To diminish the effect of color inconsistency in image alignment, we introduce to use a global color mapping (GCM) module to generate an initial sRGB image given the input raw image, which can keep the spatial location of the pixels unchanged, and the target sRGB image is utilized to guide GCM for converting the color towards it.
  Then a pre-trained optical flow estimation network (\eg, PWC-Net) is deployed to warp the target sRGB image to align with the GCM output.
  To alleviate the effect of inaccurately aligned supervision, the warped target sRGB image is leveraged to learn RAW-to-sRGB mapping.
  When training is done, the GCM module and optical flow network can be detached, thereby bringing no extra computation cost for inference.
  Experiments show that our method performs favorably against state-of-the-arts on ZRR and SR-RAW datasets.
  With our joint learning model, a light-weight backbone can achieve better quantitative and qualitative performance on ZRR dataset.
  Codes are available at \url{https://github.com/cszhilu1998/RAW-to-sRGB}.
\vspace{-5mm}
\end{abstract}
%
\section{Introduction}
%
  \begin{figure*}[t]
  \vspace{-6mm}
	\centering
	\subfigure[Input raw image (visualized)]
	{
		\begin{minipage}{.21\linewidth}
			\centering
			\includegraphics[width=\linewidth]{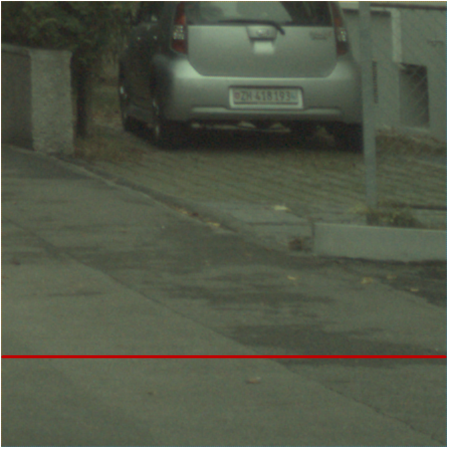}
			\vspace{-0.3cm}
		\end{minipage}
	}%
	\subfigure[PyNet~\cite{PyNet}]
	{
		\begin{minipage}{.21\linewidth}
			\centering
			\includegraphics[width=\linewidth]{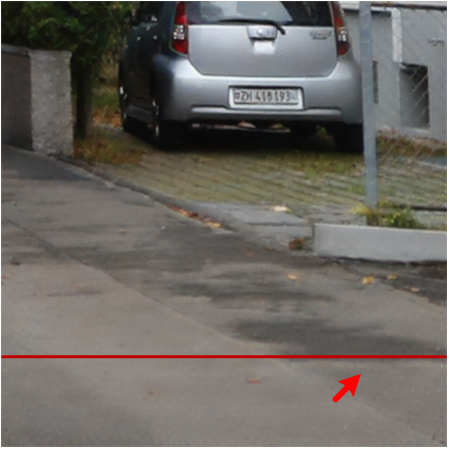}
			\vspace{-0.3cm}
		\end{minipage}
	}%
	\subfigure[Ours]
	{
		\begin{minipage}{.21\linewidth}
			\centering
			\includegraphics[width=\linewidth]{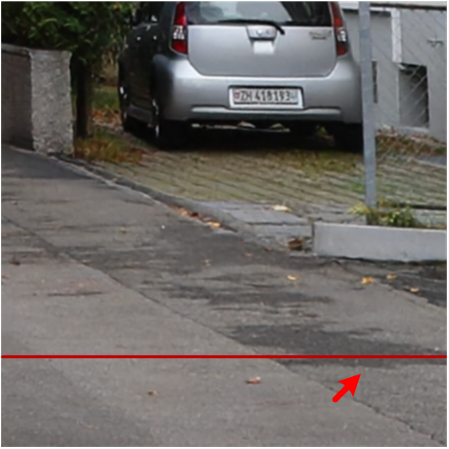}
			\vspace{-0.3cm}
		\end{minipage}
	}%
	\subfigure[Target sRGB image]
	{
		\begin{minipage}{.21\linewidth}
			\centering
			\includegraphics[width=\linewidth]{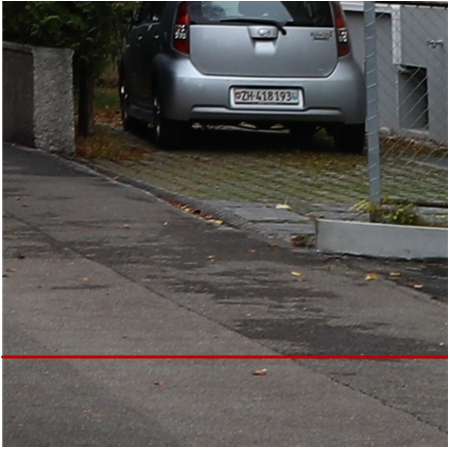}
			\vspace{-0.3cm}
		\end{minipage}
	}%

	\vspace{-3mm}
	\subfigure[Input raw image (visualized)]
	{
		\begin{minipage}{.21\linewidth}
			\centering
			\includegraphics[width=\linewidth]{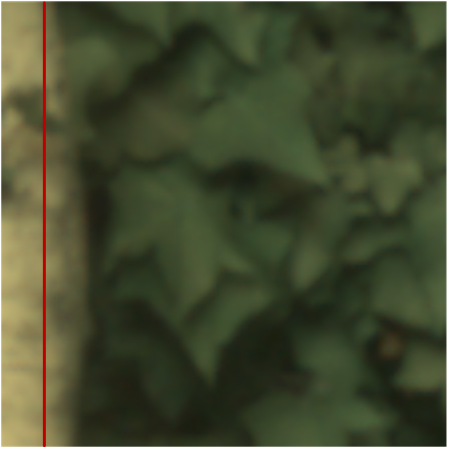}
			\vspace{-0.3cm}
		\end{minipage}
	}%
	\subfigure[Zhang \etal~\cite{SRRAW}]
	{
		\begin{minipage}{.21\linewidth}
			\centering
			\includegraphics[width=\linewidth]{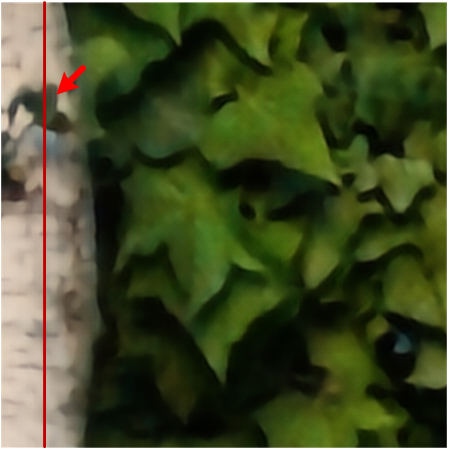}
			\vspace{-0.3cm}
		\end{minipage}
	}%
	\subfigure[Ours]
	{
		\begin{minipage}{.21\linewidth}
			\centering
			\includegraphics[width=\linewidth]{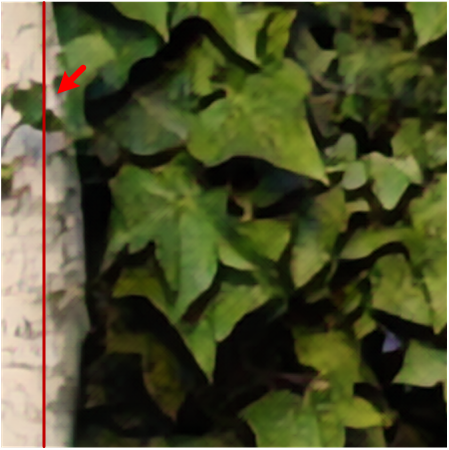}
			\vspace{-0.3cm}
		\end{minipage}
	}%
	\subfigure[Target sRGB image]
	{
		\begin{minipage}{.21\linewidth}
			\centering
			\includegraphics[width=\linewidth]{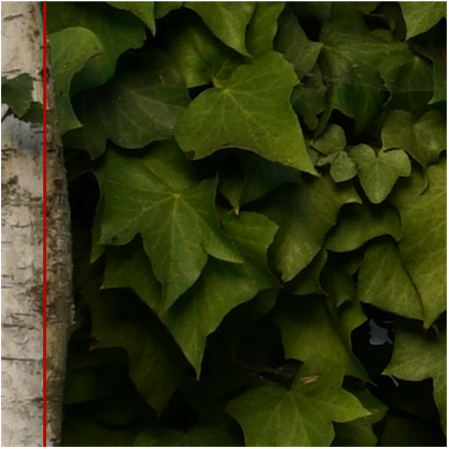}
			\vspace{-0.3cm}
		\end{minipage}
	}%
	\vspace{-4mm}
	\caption{Example of data pairs of ZRR and SR-RAW datasets, where clear spatial misalignment can be observed with the reference line. With such inaccurately aligned training data, PyNet~\cite{PyNet} and Zhang~\etal~\cite{SRRAW} are prone to generating blurry results with spatial misalignment, while our results are well aligned with the input. Please zoom in for better observation.}
	\label{fig:intro}
	\vspace{-5mm}
	\end{figure*}
  The image signal processing (ISP) pipeline refers to the processing of raw sensor image for producing high quality display-referred sRGB image, and thus is pivotal for a camera system.
  A representative ISP pipeline usually involves a sequence of steps including demosaicking, white balance, color correction, tone mapping, denoising, sharpening, gamma correction and so on~\cite{ramanath2005color}.
  While hand-crafted ISP solutions are usually adopted in current camera systems, convolutional networks (CNNs) have exhibited great potential in learning deep ISP model in an end-to-end manner~\cite{DeepISP,CameraNet,PyNet}.

  The end-to-end property of deep ISP makes it very competitive to learn RAW-to-sRGB mapping to generate high quality image for mobile camera~\cite{PyNet}.
  Albeit mobile camera has become the dominant sources of photos, it has a smaller sensor size and limited aperture in comparison to DSLR camera.
  By learning RAW-to-sRGB mapping to produce DSLR-like sRGB image from mobile raw image, deep ISP model can thus offer an encouraging way to close the gap between mobile camera and DSLR camera.
  Moreover, in contrast to 8-bit sRGB image, raw image usually has higher-bit (\eg, 10-14 bit) and may convey richer details.
  Therefore, learning RAW-to-sRGB mapping is beneficial to performance improvement even for other low level vision tasks, \eg, image super-resolution~\cite{SRRAW}, low light image denoising~\cite{SID} and high dynamic range imaging (HDR)~\cite{MergingISP}.

  However, when preparing training data, input raw image and target sRGB image are usually taken using different cameras (\eg, a smartphone and a DSLR) or with different camera configurations (\eg, focal length).
  Consequently, color inconsistency and spatial misalignment are usually inevitable.
  On the one hand, the color inconsistency makes it very challenging to generate well-aligned training pairs of input raw and target sRGB images.
  The input raw and target sRGB images usually cannot be perfectly aligned by existing methods~\cite{SIFT,RANSAC}, resulting in mild alignment.
  On the other hand, learning with inaccurately aligned supervision is prone to pixel shift and producing blurry results (see Fig.~\ref{fig:intro}(b)).
  To alleviate the adverse effect of inaccurate alignment, AWNet~\cite{AWNet} adopted global context block~\cite{GCNet} at the cost of increasing inference time, while Zhang \etal~\cite{SRRAW} presented a contextual bilateral (CoBi) loss to search the best matching patch for supervision.
  However, the patch-based alignment is unable to appropriately handle the spatially variant misalignment caused by depth discrepancy between objects.
  As a result, their method is still prone to producing blurry results as shown in Fig.~\ref{fig:intro}(f).

  In order to circumvent inaccurately aligned supervision problem, this paper presents a joint learning model for image alignment and RAW-to-sRGB mapping.
  We argue that one major reason that explains the inaccurate/mild alignment is the severe color inconsistency between input raw and target sRGB images.
  Otherwise, existing optical flow networks~\cite{FlowNet,FlowNet_v2,PWC-Net} can be readily utilized to fulfill the task of image alignment.
  Thus, we suggest to perform image alignment by concatenating a delicately designed global color mapping (GCM) module with a pre-trained optical flow estimation network (\eg, PWC-Net~\cite{PWC-Net}).
  In particular, the GCM module involves a stack of $1 \times 1$ convolutional layers to ensure that the mapping is spatially independent.
  To overcome the obstacle of color inconsistency, we constrain the GCM output to approximate the aligned target sRGB image.
  It is worth noting that GCM is deployed to align target sRGB image only during training.
  Thus, we can also take the target sRGB image and coordinate map to generate conditional guidance for modulating GCM features towards diminishing color inconsistency.
  Then, a pre-trained optical flow estimation network (\eg, PWC-Net~\cite{PWC-Net}) can be used to align the target sRGB image with the GCM output, resulting in the well aligned sRGB image.

  The aligned target sRGB image can serve as a better supervision for training the RAW-to-sRGB mapping.
  In particular, we propose a LiteISPNet by reducing the residual channel attention blocks (RCABs) in MW-ISPNet~\cite{AIM2020ISP}.
  GCM and LiteISPNet are jointly trained for both the alignment of target sRGB image (\ie, GCM and PWC-Net) and the RAW-to-sRGB mapping (\ie, LiteISPNet).
  When training is done, GCM and PWC-Net can be detached and only LiteISPNet is required for handling test raw images, thereby bringing no extra inference cost.
  Experiments on Zurich RAW to RGB (ZRR) dataset~\cite{PyNet} show that our solution is effective in learning with inaccurately aligned supervision and producing more fine details.
  Our proposed method also outperforms the state-of-the-art method in terms of quantitative metrics, perceptual quality and computational efficiency.
  Furthermore, using SRResNet as the backbone, experiments also show the effectiveness of our method for image super-resolution on the SR-RAW dataset~\cite{SRRAW}.

  The main contributions of this work are three-fold:
  \begin{itemize}
    \vspace{-1.5mm}  
    \item An effective approach is presented to circumvent the task of learning RAW-to-sRGB mapping with inaccurately aligned supervision.
    \vspace{-2mm}
    \item A global color mapping (GCM) module is delicately designed to tackle the effect of color inconsistency on image alignment.
          A spatially preserving network (SPN) is leveraged to avoid spatial shift of pixels, and the target sRGB image is adopted to modulate GCM features towards diminishing color inconsistency.
    \vspace{-2mm}
    \item Quantitative and qualitative results on ZRR and SR-RAW datasets show that our method outperforms the state-of-the-art methods with no extra inference cost.
    \vspace{-1.5mm}  
  \end{itemize}
%
\vspace{-2mm}
\section{Related Work}\label{sec:RelatedWork}
%
\vspace{-1mm}
\subsection{Deep Networks for ISP}\label{sec:2.1}
\vspace{-1mm}
  The camera ISP pipeline is deployed to produce display-referred sRGB image from raw images.
  To this end, classical ISP has been extensively studied, which generally involves a sequence of subtasks~\cite{ramanath2005color} including demosaicking, white balance, color correction, tone mapping, denoising, sharpening, gamma correction and \etc.
  For each subtask, a number of methods have been proposed in the literature~\cite{buades2005non,hirakawa2005adaptive,van2007edge,rizzi2003new}.
  Motivated by the unprecedented success of deep learning, CNNs have also been investigated to tackle several hard ISP subtasks like image denoising \cite{zhang2017beyond,zhang2018ffdnet,CycleISP}, demosaicking~\cite{gharbi2016deep,tan2017color,liu2020joint}, auto-white-balance (AWB)~\cite{lou2015color,barron2015convolutional,hu2017fc4,xiao2020multi} and tone mapping~\cite{zeng2020learning,CSRNet,gharbi2017deep,cai2018learning}.

  Recently, several attempts have been made to learn a full ISP pipeline with a deep model.
  Schwartz~\etal~\cite{DeepISP} designed a DeepISP model to produce a well-lit sRGB image given a raw image captured under low-light environment.
  CameraNet~\cite{CameraNet} explicitly grouped the sub-tasks into two categories (\ie, restoration and enhancement), and extracted the ground-truths by commercial software.
  Ignatov~\etal~\cite{PyNet} collected a dataset containing paired raw and sRGB images, which respectively are captured by Huawei P20 smartphone and Canon 5D Mark IV DSLR. 
  With the dataset proposed in~\cite{PyNet}, two challenges were held~\cite{AIM2019ISP,AIM2020ISP}.
  Among the decent methods proposed by the participants, MW-ISPNet~\cite{AIM2020ISP} leveraged MWCNN~\cite{MWCNN} and residual channel attention blocks (RCABs)~\cite{RCAN}, AWNet~\cite{AWNet} adopted global context block~\cite{GCNet} to learn non-local color mapping, and they won first two places in the perceptual track~\cite{AIM2020ISP}.
  In this work, we present a LiteISPNet by reducing the number of RCABs in MW-ISPNet~\cite{AIM2020ISP} for learning full ISP model.
  By incorporating LiteISPNet with the joint learning model, better quantitative results and perceptual quality can be attained in comparison to MW-ISPNet~\cite{AIM2020ISP} and AWNet~\cite{AWNet}.

\vspace{-1mm}
\subsection{RAW-to-sRGB Mappings for More Tasks}\label{sec:2.2}
\vspace{-1mm}
  In~\cite{PyNet}, the raw sensor and sRGB images are captured using different cameras.
  This makes it feasible to empower low-quality raw sensor to produce high-quality sRGB image by imitating either other cameras or camera with different configurations.
  Moreover, the pixels of raw images are usually of higher-bit (\eg, $10 \sim 14$ bit), spatially independent and linear to brightness, thereby conveying richer details for benefiting image enhancement.
  Chen~\etal~\cite{SID} pioneered this line of work by leveraging paired low-light raw and long-exposure sRGB images with different ISO settings for extreme low-light image enhancement.
  And they further explored extreme low-light video enhancement in \cite{SIDMotion}.
  Analogously, raw images are also utilized in other low level vision tasks such as HDR~\cite{MergingISP} and image super-resolution (SR)~\cite{SRRAW,xu2019towards}.
  In this work, our method is also tested on image SR~\cite{SRRAW} by using SRResNet~\cite{SRResNet} as backbone, and achieves better quantitative and qualitative results.

  Furthermore, studies have also been given for repurposing or merging RAW-to-sRGB mapping with high-level vision tasks.
  Wu~\etal~\cite{VisionISP} designed a visionISP model to generate better input for object detection.
  Schwartz~\etal~\cite{ISPDistillation} learned a model for image classification with raw images via distilling the knowledge of an ISP pipeline and an sRGB image classification model.

\vspace{-1mm}
\subsection{Alignment of Paired Raw and sRGB Images}
\vspace{-1mm}
  For learning RAW-to-sRGB mappings, input raw and target sRGB images are usually taken using different cameras or with different camera configurations~\cite{PyNet,SRRAW}.
  Misalignment caused by multiple cameras and motion in the scene are thus inevitable, hindering the learning of RAW-to-sRGB mappings and giving rise to blurry or even pixel-shifted results.
  For suppressing the effect of motion in the scene, dual or multiple cameras are deployed to shoot concurrently for some datasets like KITTI~\cite{KITTI2012,KITTI2015} and MultiPIE~\cite{MultiPIE}.
  Beam splitter is also introduced to collect image pairs at the ``same position'' with different settings for real-world super-resolution~\cite{joze2020imagepairs} and deblurring~\cite{rim2020real}.
  Albeit with such equipments, misalignment remains unavoidable.

  Several methods have been presented to align images from different sources.
  SIFT keypoints~\cite{SIFT} are adopted for image registration in \cite{ignatov2017dslr,PyNet,wei2020component}, where the homography can be estimated via RANSAC algorithm~\cite{RANSAC}.
  Cai~\etal~\cite{cai2019toward} designed a pixel-wise registration method that considers luminance adjustment for a real-world super-resolution dataset.
  Li~\etal~\cite{li2018learning} warped the guidance image for facial image restoration via optical flow based methods~\cite{FlowNet,FlowNet_v2,PWC-Net}.
  However, input raw and target sRGB images usually have severe color inconsistency and cannot be perfectly aligned by existing methods, thereby resulting in mild alignment.
  Global context block~\cite{GCNet,AWNet} and CoBi loss~\cite{SRRAW} have been introduced to alleviate the effect of mild alignment, but are still prone to producing blurry results.
  In this work, we introduce a global color mapping (GCM) module for tackling color inconsistency, and present a joint learning model for both image alignment and RAW-to-sRGB mapping.

\vspace{-1.5mm}
\section{Proposed Method}\label{sec:Method}
\vspace{-1.5mm}
%
In this section, we first give an overview of our model for joint learning of image alignment and RAW-to-sRGB mapping.
Then, the design of the global color mapping (GCM) module and LiteISPNet is detailed, and the learning objectives are presented.
And we also extend our proposed method to other image enhancement tasks, \eg, image SR.

\vspace{-1mm}
\subsection{Joint Learning Model}
\vspace{-1mm}

\begin{figure*}[h]
	\centering
	\vspace{-2mm}
	\begin{overpic}[width=.9\linewidth]{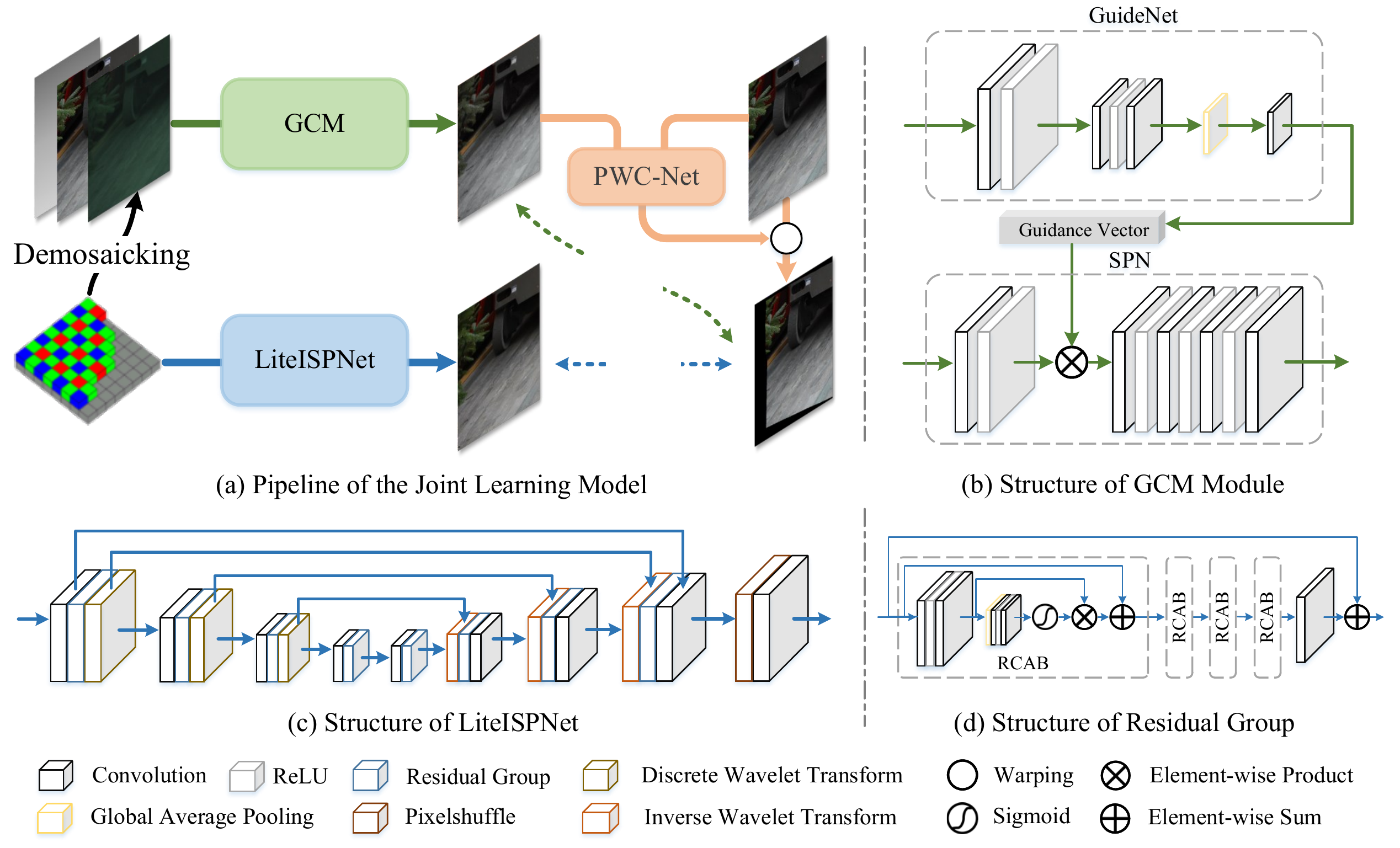}

		\put(4.5, 59){$\mathbf{\bm{\tau}}\ \ \ \mathbf{y}\ \ \ \hat{\mathbf{x}}$}
		\put(34.9,58.3){$\tilde{\mathbf{y}}$}
		\put(55.5,58.3){$\mathbf{y}$}
		
		\put(41.8,41.1){\footnotesize $\mathcal{L}_\mathrm{GCM}$}
		\put(41.5,39.1){\footnotesize Eqn.~(\ref{eqn:loss_GCM})}
		\put(44.1,34.8){\footnotesize $\mathcal{L}_\mathrm{ISP}$}
    \put(43.4,33.0){\footnotesize Eqn.~(\ref{eqn:loss_isp})}
    
    \put(55.3,42.9){\scriptsize $\mathcal{W}$}
    
		\put(5.6,27.8){$\mathbf{x}$}
		\put(34.9,27.8){$\hat{\mathbf{y}}$}
		\put(55.5,27.8){$\mathbf{y}^\mathit{w}$}

		\put(62.7,50.9){$\begin{matrix}\bm{\tau}\\\mathbf{y}\vspace{0.7mm}\\\hat{\mathbf{x}}\end{matrix}$}
		\put(62.7,33.8){$\begin{matrix}\bm{\tau}\\\hat{\mathbf{x}}\end{matrix}$}
		\put(96.9,33.8){$\tilde{\mathbf{y}}$}
		
		\put(-0.2,15.8){$\mathbf{x}$}
		\put(59.5,15.8){$\hat{\mathbf{y}}$}
		
		\put(67.8,4.5){\scriptsize $\mathcal{W}$}
		
	\end{overpic}
	\vspace{-3mm}
	\caption{Illustration of the proposed joint learning framework.
		(a)~Pipeline of the joint learning model, where GCM module converts the color of $\hat{\mathbf{x}}$ (demosaicked from $\mathbf{x}$ via simple interpolation) for more accurate optical flow, then the warped sRGB ($\mathbf{y}^\mathit{w}$) can provide aligned supervision for a joint training of GCM and LiteISPNet.
		(b)~Structure of GCM module, which is composed of a GuideNet and an SPN. The target sRGB ($\mathbf{y}$) and a coordinate map ($\bm{\tau}$) are deployed for guiding the color conversion.
		(c)~Structure of LiteISPNet, please refer to Sec.~\ref{sec:LiteISPNet} for more details.
		(d)~Structure of residual group used in (c).}
	\label{fig:framework}
	\vspace{-0.5cm}
\end{figure*}
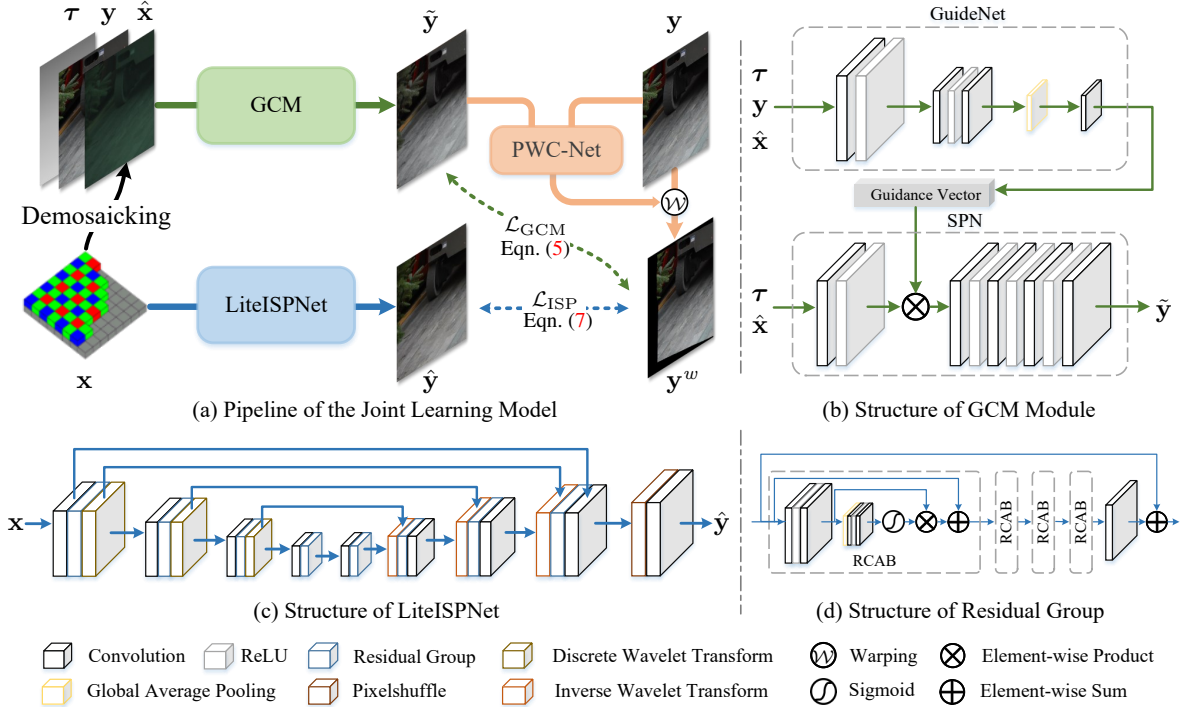

  Denote by $\mathbf{x}$ and $\mathbf{y}$ a raw image and the corresponding target sRGB image.
  The RAW-to-sRGB mapping is used to produce a sRGB image $\hat{\mathbf{y}}$ from $\mathbf{x}$ for approximating the color characteristic of target sRGB image $\mathbf{y}$,
    \vspace{-1mm}  
    \begin{equation}
      \hat{\mathbf{y}} = \mathcal{I}(\mathbf{x}; \Theta_\mathcal{I}),
    \label{eqn:ISP}
    \vspace{-1mm}  
    \end{equation}
  where $\mathcal{I}$ denotes a RAW-to-sRGB mapping (\eg, LiteISPNet in Sec.~\ref{sec:LiteISPNet}) with the parameter $\Theta_\mathcal{I}$.

  However, $\mathbf{x}$ and $\mathbf{y}$ are usually taken using different cameras or with diverse camera configurations, giving rise to inevitable spatial misalignment between them.
  Moreover, the severe color inconsistency between $\mathbf{x}$ and $\mathbf{y}$ further makes the image alignment more difficult.
  On the other hand, the RAW-to-sRGB mapping aims at imitating the color characteristics and fine details of target sRGB image.
  The misalignment between $\mathbf{x}$ and $\mathbf{y}$ is harmful to the learning of RAW-to-sRGB mapping, thereby being prone to producing blurry outputs with unfavorable pixel shift.
  Several approaches have been proposed to improve the alignment or the learning robustness, but are still not sufficient in suppressing blurry outputs.

  Joint learning of image alignment and RAW-to-sRGB mapping can offer some new chances to circumvent such issue.
  On the one hand, RAW-to-sRGB mapping is helpful in diminishing the color inconsistency between $\mathbf{x}$ and $\mathbf{y}$, thus easing the difficulty of image alignment.
  On the other hand, better image alignment also benefits RAW-to-sRGB mapping for suppressing blurry outputs and pixel shift.
  Unfortunately, the RAW-to-sRGB mapping (\eg, \cite{PyNet}) cannot completely avoid pixel shift (see Fig.~\ref{fig:intro}), thus aligning $\mathbf{y}$ with $\hat{\mathbf{y}}$ cannot solve the misalignment problem.

  Instead of aligning $\mathbf{y}$ with $\hat{\mathbf{y}}$, our joint learning model leverages a delicately designed global color mapping (GCM) module to generate a color-adjusted image $\tilde{\mathbf{y}}$ for warping $\mathbf{y}$.
  A simple demosaicking method (\eg, bicubic) is first used to obtain $\hat{\mathbf{x}}$ by filling in missing values of $\mathbf{x}$.
  Then, the GCM module is introduced as a pixel-wise mapping of $\hat{\mathbf{x}}$, and thus can guarantee not to introduce any pixel shift in color correction.
  Furthermore, GCM is only required during training.
  So we can take target sRGB image and coordinate map as input to generate conditional guidance for modulating GCM features towards diminishing color inconsistency.
  Thus, the GCM module can be given by,
    \vspace{-2.5mm}
    \begin{equation}
    \tilde{\mathbf{y}} = \mathcal{C}(\hat{\mathbf{x}}, \mathbf{y}, \bm{\tau}; \Theta_\mathcal{C}),
    \label{eqn:GCM}
    \vspace{-1mm}
    \end{equation}
  where $\mathcal{C}$ denotes the GCM module, $\bm{\tau}\in\mathbb{R}^{2\times H \times W}$ is the 2D coordinate map containing the coordinate of the pixels, which is normalized to $[-1, 1]$.
  Given $\tilde{\mathbf{y}}$ and $\mathbf{y}$, we use a pre-trained optical flow network (denote by $\mathcal{F}$), \eg, PWC-Net~\cite{PWC-Net}, to estimate the optical flow $\Psi$,
    \vspace{-2mm}  
    \begin{equation}
      \Psi = \mathcal{F}(\tilde{\mathbf{y}}, \mathbf{y}).
    \label{eqn:flow}
    \vspace{-1mm}  
    \end{equation}
  The estimated optical flow is then used to warp $\mathbf{y}$ to form a warped target sRGB image,
    \vspace{-1.5mm}  
    \begin{equation}
      \mathbf{y}^\mathit{w} = \mathcal{W}(\mathbf{y}, \Psi),
      \label{eqn:warp_srgb}
    \vspace{-1mm}  
    \end{equation}
  where $\mathcal{W}$ is a warping operation (\eg, bilinear interpolation)~\cite{PWC-Net}.
  Then, $\mathbf{y}^\mathit{w}$ can serve as a well aligned target sRGB image for supervising the RAW-to-sRGB mapping $\mathcal{I}$ in Eqn.~(\ref{eqn:ISP}), resulting in our joint learning model (see Fig.~\ref{fig:framework}).

\vspace{-1mm}
\subsection{GCM Module}
\vspace{-1mm}

  For optical flow estimation, the color-adjusted image $\tilde{\mathbf{y}}$  is required to satisfy two prerequisites.
  (i)~$\tilde{\mathbf{y}}$ should imitate the color of $\mathbf{y}$ for diminishing the severe color inconsistency.
  (ii)~The spatial position of the pixels should keep the same as the input image $\hat{\mathbf{x}}$.
  According to \cite{CSRNet}, some commonly-used image processing operations can be approximated or formulated by multi-layer perceptron (MLP), and the pixel-wise nature ensures that the input and output are strictly aligned.
  Thus, we deploy a spatially preserving network (SPN) as the backbone of our GCM module, which is composed of a stack of $1 \times 1$ convolutional layers.

  It is worth noting that GCM is only required during training.
  Thus the target sRGB image can also be safely fed into GCM for better converting $\tilde{\mathbf{y}}$ towards target sRGB image.
  To this end, we build a GuideNet to generate a conditional guidance vector from the raw ($\hat{\mathbf{x}}$) and target sRGB ($\mathbf{y}$) pair (see Fig.~\ref{fig:framework}(b)).
  To alleviate the effect of misalignment in generating guidance vector, we use a relatively large kernel (\ie, $7 \times 7$) in GuideNet, and global average pooling is deployed to acquire holistic information.

  Besides, dark corner (\emph{a.k.a}, vignetting) can be observed in the raw images of several datasets (\eg, ZRR~\cite{AIM2020ISP}), where the brightness is gradually weakened from image center to the boarders.
  Standard global color mapping, however, is not sufficient for compensating the adverse effect of dark corner.
  Fortunately, the effect of dark corner can be modeled by a pixel-wise function of 2D coordinate map (\ie, $\bm{\tau}$ in Eqn.~(\ref{eqn:GCM}))~\cite{yu2004practical}.
  As a remedy, we simply feed $\bm{\tau}$ to both the SPN and the GuideNet for handling anti-vignetting while learning color mapping simultaneously.

  With the GCM output $\tilde{\mathbf{y}}$, we use PWC-Net~\cite{PWC-Net} to estimate the optical flow for warping target sRGB image $\mathbf{y}$.
  The warped sRGB image $\mathbf{y}^{w}$ can then be adopted as the supervision for training GCM.
  Besides, we note that the pixel-wise mapping makes GCM unable to remove the noise in $\hat{\mathbf{x}}$.
  Nonetheless, benefiting from PWC-Net~\cite{PWC-Net}, we can still estimate the optical flow between $\tilde{\mathbf{y}}$ and $\mathbf{y}$ robustly.

\vspace{-1mm}
\subsection{LiteISPNet}\label{sec:LiteISPNet}
\vspace{-1mm}
  For alleviating mild alignment, existing methods usually build large models and exploit specific modules~\cite{PyNet,AWNet}, which improves the performance at the cost of increasing inference time.
  Considering that a better alignment can be attained by joint learning, we can adopt a more efficient network for learning RAW-to-sRGB mapping to achieve comparable or even better performance.
  Thus, we present a LiteISPNet by simplifying MW-ISPNet~\cite{AIM2020ISP}, which is a U-Net~\cite{U-Net} based multi-level wavelet ISP network.
  In particular, we put convolutional layer and residual group~\cite{RCAN} before each wavelet decomposition by referring to~\cite{liu2019multi}.
  Moreover, we also reduce the number of RCAB from 20 to 4 in each residual group to construct the LiteISPNet backbone.
  Fig.~\ref{fig:framework}(c) illustrates the network structure of LiteISPNet. 
  Benefited from the structure modification and joint learning, LiteISPNet outperforms MW-ISPNet~\cite{AIM2020ISP} quantitatively and qualitatively with $\sim$40\% model size and $\sim$20\% running time.

\vspace{-1mm}
\subsection{Learning Objectives}
\vspace{-1mm}
  Using the pre-trained PWC-Net~\cite{PWC-Net} for computing optical flow, GCM and LiteISPNet can be jointly trained for learning image alignment and RAW-to-sRGB mapping.
  In the following, we respectively introduce the loss terms for GCM and LiteISPNet.

  \noindent\textbf{Loss Term for GCM}.
  Denote by $\tilde{\mathbf{y}}$ the GCM output in Eqn.~\ref{eqn:GCM} and $\mathbf{y}^w$ the warped target sRGB image in Eqn.~\ref{eqn:warp_srgb}.
  The loss term for GCM is given by,
    \vspace{-1mm}  
     \begin{equation}
      \mathcal{L}_\mathrm{GCM}(\tilde{\mathbf{y}}, \mathbf{y}^w) = \| \mathbf{m} \circ (\tilde{\mathbf{y}} - \mathbf{y}^w) \|_1,
    \label{eqn:loss_GCM}
    \vspace{-1mm}  
    \end{equation}
  where $\circ$ denotes entry-wise product, $\|\cdot\|_1$ is $\ell_1$ loss and $\mathbf{m}$ is a mask indicating valid positions of the optical flow.
  Here, each element ${m}_i$ of $\mathbf{m}$ is defined as,
    \vspace{-1mm}  
      \begin{equation}
      {m}_i = \left\{
        \begin{array}{cl}
          1,\ & [\mathcal{W}(\mathbf{1},\Psi)]_i \ge 1-\epsilon\\
          0,\ & \text{otherwise}
        \end{array}\right.,
    \vspace{-1mm}  
    \end{equation}
  where $\mathbf{1}$ denotes an all-1 matrix, $\epsilon$ is a threshold set to 0.001, and $[\cdot]_i$ denotes the $i$-th element of a matrix.

  \noindent\textbf{Loss Terms for LiteISPNet}.
  Denote by $\hat{\mathbf{y}}$ the LiteISPNet output in Eqn.~\ref{eqn:ISP}.
  The LiteISPNet is trained with a combination of $\ell_1$ loss and (VGG-based) perceptual loss~\cite{VGGLOSS}, which can be written as,
    \vspace{-1mm}  
  \begin{equation}
    \begin{split}
    \mathcal{L}_\mathrm{ISP}(\hat{\mathbf{y}}, \mathbf{y}^\mathit{w})
    & = \lambda_{\ell_1} \|\mathbf{m} \circ (\hat{\mathbf{y}} - \mathbf{y}^\mathit{w})\|_1 \\
    & + \lambda_\mathit{VGG} \|\mathbf{m} \circ (\phi(\hat{\mathbf{y}}) - \phi(\mathbf{y}^\mathit{w}))\|_1,
    \end{split}
    \label{eqn:loss_isp}
    \vspace{-1mm}  
  \end{equation}
  where $\phi$ denotes the pre-trained VGG-19~\cite{VGGLOSS} network, and we set $\lambda_{\ell_1}=\lambda_\mathit{VGG}=1$.
  Besides, to further enhance the visual quality, we also train the LiteISPNet with adversarial loss~\cite{GAN}. Following LSGAN~\cite{LSGAN}, the loss function is defined as,
    \vspace{-1mm}  
    \begin{equation}
      \mathcal{L}_\mathit{GAN} = \frac{1}{2}\mathbb{E}_{\mathbf{x}\sim\mathit{p}_\mathbf{x}}[\mathcal{D}(\mathcal{I}(\mathbf{x})) - 1]^2,
    \vspace{-1mm}  
    \end{equation}
  where $\mathcal{D}$ denotes the discriminator (see the suppl for detailed structure configuration), which is trained by
    \vspace{-1mm}  
    \begin{equation}
      \mathcal{L}_\mathit{D} = \frac{1}{2}\mathbb{E}_{\mathbf{y}\sim\mathit{p}_\mathbf{y}}[\mathcal{D}(\mathbf{y}) - 1]^2 +
                               \frac{1}{2}\mathbb{E}_{\mathbf{x}\sim\mathit{p}_\mathbf{x}}[\mathcal{D}(\mathcal{I}(\mathbf{x}))]^2.
    \vspace{-1mm}  
    \end{equation}
  Then, LiteISPGAN is provided by training with the loss,
    \vspace{-1mm}  
    \begin{equation}
      \mathcal{L}_\mathrm{ISPGAN} = \mathcal{L}_\mathrm{ISP} + \lambda_\mathit{GAN} \mathcal{L}_\mathit{GAN},
    \vspace{-1mm}  
    \end{equation}
    where $\lambda_\mathit{GAN}=0.01$.

  \noindent\textbf{Learning Objective}.
  With the above loss terms, the overall learning objective of our model can be defined by,
    \vspace{-1mm}  
    \begin{equation}
      \mathcal{L} = \mathcal{L}_\mathrm{GCM} + \mathcal{L}_\mathrm{ISP/ISPGAN}.
    \vspace{-1mm}  
    \end{equation}

  \begin{figure*}[t] 
    \vspace{-6mm}  
    \centering
    \subfigure[Raw image (visualized)]
    {
      \begin{minipage}{.21\linewidth}
        \centering
        \includegraphics[width=\linewidth]{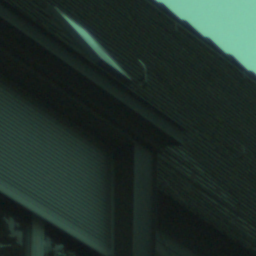}
        \vspace{-0.3cm}
      \end{minipage}
    }%
    \subfigure[PyNet~\cite{PyNet}]
    {
      \begin{minipage}{.21\linewidth}
        \centering
        \includegraphics[width=\linewidth]{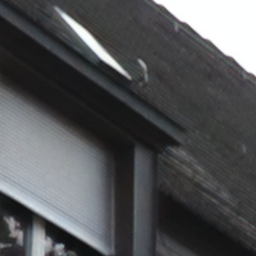}
        \vspace{-0.3cm}
      \end{minipage}
    }%
    \subfigure[AWNet (raw)~\cite{AWNet}]
    {
      \begin{minipage}{.21\linewidth}
        \centering
        \includegraphics[width=\linewidth]{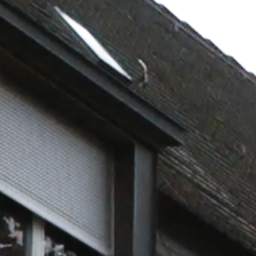}
        \vspace{-0.3cm}
      \end{minipage}
    }%
    \subfigure[AWNet (demosaicked)~\cite{AWNet}]
    {
      \begin{minipage}{.21\linewidth}
        \centering
        \includegraphics[width=\linewidth]{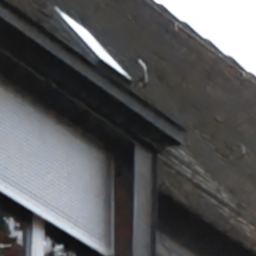}
        \vspace{-0.3cm}
      \end{minipage}
    }%
  
  	\vspace{-3mm}
    \subfigure[MW-ISPNet (GAN)~\cite{AIM2020ISP}]
    {
      \begin{minipage}{.21\linewidth}
        \centering
        \includegraphics[width=\linewidth]{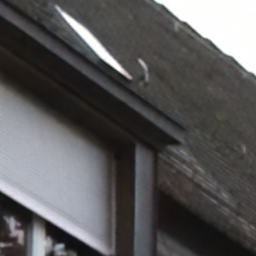}
        \vspace{-0.3cm}
      \end{minipage}
    }%
    \subfigure[Ours (LiteISPNet)]
    {
      \begin{minipage}{.21\linewidth}
        \centering
        \includegraphics[width=\linewidth]{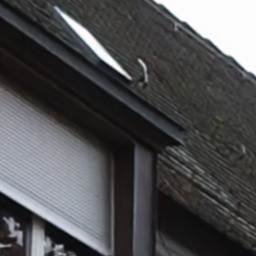}
        \vspace{-0.3cm}
      \end{minipage}
    }%
    \subfigure[Ours (LiteISPGAN)]
    {
      \begin{minipage}{.21\linewidth}
        \centering
        \includegraphics[width=\linewidth]{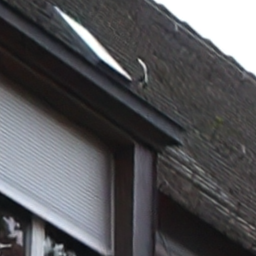}
        \vspace{-0.3cm}
      \end{minipage}
    }%
    \subfigure[GT]
    {
      \begin{minipage}{.21\linewidth}
        \centering
        \includegraphics[width=\linewidth]{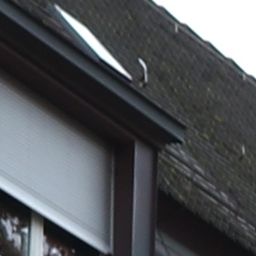}
        \vspace{-0.3cm}
      \end{minipage}
    }%
    \vspace{-4mm}
    \caption{Visual comparisons on ZRR dataset. Please zoom in for better observation.}
    \label{fig:ZRR-results1}
    \vspace{-3mm}
  \end{figure*}    

\begin{table*}[t] 
  \small
  \renewcommand\arraystretch{1}
  \begin{center}
    \caption{Quantitative results on ZRR dataset for methods trained without and with adversarial loss.}\label{tab:ZRR-results}
    \vspace{-2mm}
    \scalebox{0.9}{
    \begin{tabular}{cccccc}
      \toprule
      Method & \tabincell{c}{\# Params\\ {(M)}}  & \tabincell{c}{Time \\ {(ms)}}
      & \tabincell{c}{\emph{Original GT} \\ \footnotesize{PSNR}\scriptsize{$\uparrow$} \small{/} \footnotesize{SSIM}\scriptsize{$\uparrow$} \small{/} \footnotesize{LPIPS}\scriptsize{$\downarrow$}}
      & \tabincell{c}{\emph{Align GT with raw} \\ \footnotesize{PSNR}\scriptsize{$\uparrow$} \small{/} \footnotesize{SSIM}\scriptsize{$\uparrow$} \small{/} \footnotesize{LPIPS}\scriptsize{$\downarrow$}}
      & \tabincell{c}{\emph{Align GT with result} \\ \footnotesize{PSNR}\scriptsize{$\uparrow$} \small{/} \footnotesize{SSIM}\scriptsize{$\uparrow$} \small{/} \footnotesize{LPIPS}\scriptsize{$\downarrow$}} \\

      \midrule
      PyNet~\cite{PyNet}                & 47.6 & 62.7     & 21.19 / 0.7471 / 0.193    & 22.73 / 0.8451 / 0.152 & 22.97 / 0.8510 / 0.152 \\
      AWNet (raw)~\cite{AWNet}          & 52.2 & 55.7     & 21.42 / 0.7478 / 0.198    & 23.27 / 0.8542 / 0.151 & 23.35 / 0.8559 / 0.151   \\
      AWNet (demosaicked)~\cite{AWNet}  & 50.1 & 62.7     & 21.53 / 0.7488 / 0.212    & 23.38 / 0.8497 / 0.164 & 23.41 / 0.8502 / 0.164    \\
      MW-ISPNet~\cite{AIM2020ISP}       & 29.2 & 110.5    & 21.42 / \textbf{0.7544} / 0.213    & 23.07 / 0.8479 / 0.165 & 23.31 / 0.8578 / 0.164   \\
      Ours (LiteISPNet)       & 11.9  & 23.3
      & \textbf{21.55} / 0.7487 / \textbf{0.187}
      & \textbf{23.76} / \textbf{0.8730} / \textbf{0.133}
      & \textbf{23.87} / \textbf{0.8737} / \textbf{0.133} \\
      \hline
      MW-ISPNet (GAN)~\cite{AIM2020ISP}     & 29.2 & 110.5   & 21.16 / 0.7317 / \textbf{0.159}    & 22.80 / 0.8285 / 0.134 & 23.38 / 0.8513 / 0.131   \\
      Ours (LiteISPGAN)       & 11.9 &  23.3
      & \textbf{21.28} / \textbf{0.7387} / \textbf{0.159}
      & \textbf{23.47} / \textbf{0.8642} / \textbf{0.120}
      & \textbf{23.56} / \textbf{0.8670} / \textbf{0.119}\\
      \bottomrule
    \end{tabular}}
  \end{center}
  \vspace{-8mm}
\end{table*}

\vspace{-1mm}
\subsection{Extension to Other Image Enhancement Tasks}
\vspace{-1mm}
  As previously discussed in Sec.~\ref{sec:2.2}, RAW-to-sRGB mapping has been combined with many other tasks, where considerable efforts have been paid to mitigate the influence of misalignment~\cite{SRRAW,xu2019towards,MergingISP}.
  In these scenarios, the main obstacle to alignment is similar with the ISP problem~\cite{PyNet}, thus the proposed joint training framework can be naturally extended to many image enhancement tasks.
  In this paper, we conduct experiments on the raw image SR~\cite{SRRAW} task to show the generalization ability of our method.

  \begin{figure*}[t] 
    \vspace{-6mm}  
    \centering
    \subfigure[Bicubic$^\dagger$]
    {
     \begin{minipage}{.18\linewidth}
        \centering
        \includegraphics[width=\linewidth]{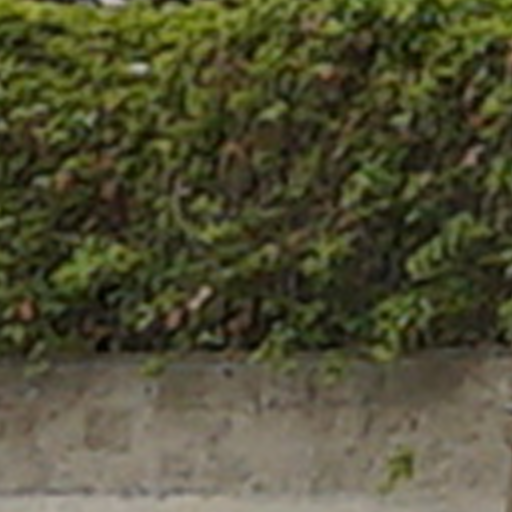}
        \vspace{-0.3cm}
     \end{minipage}
    }%
    \subfigure[SRGAN$^\dagger$~\cite{SRResNet}]
    {
     \begin{minipage}{.18\linewidth}
        \centering
        \includegraphics[width=\linewidth]{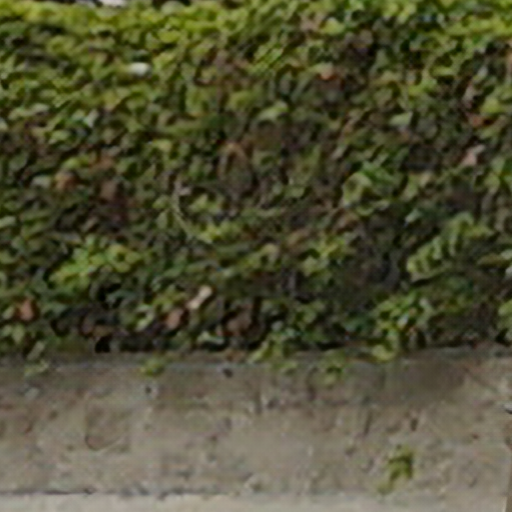}
        \vspace{-0.3cm}
     \end{minipage}
    }%
    \subfigure[ESRGAN$^\dagger$~\cite{ESRGAN}]
    {
     \begin{minipage}{.18\linewidth}
        \centering
        \includegraphics[width=\linewidth]{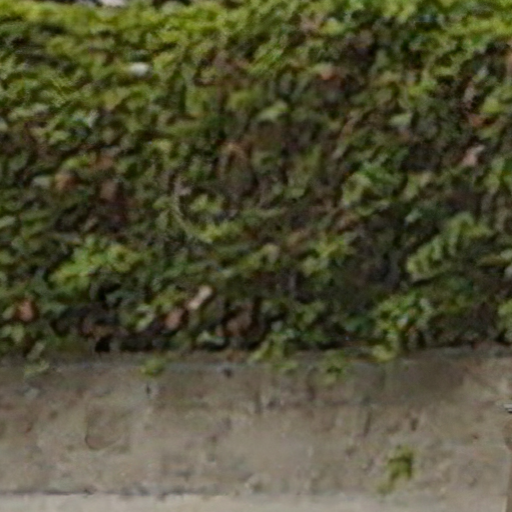}
        \vspace{-0.3cm}
     \end{minipage}
    }%
    \subfigure[SPSR$^\dagger$~\cite{SPSR}]
    {
     \begin{minipage}{.18\linewidth}
        \centering
        \includegraphics[width=\linewidth]{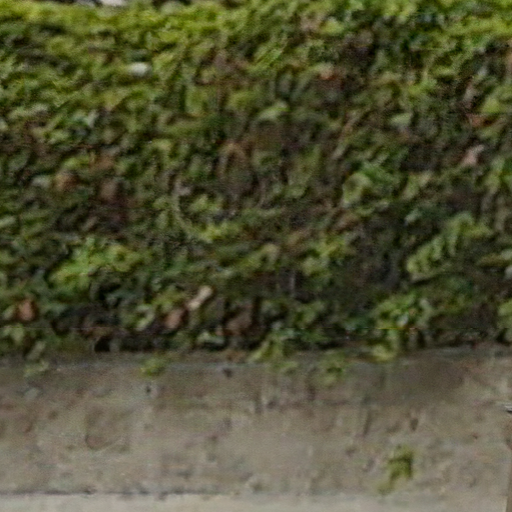}
        \vspace{-0.3cm}
     \end{minipage}
    }%
    \subfigure[RealSR$^\dagger$~\cite{RealSR}]
    {
     \begin{minipage}{.18\linewidth}
        \centering
        \includegraphics[width=\linewidth]{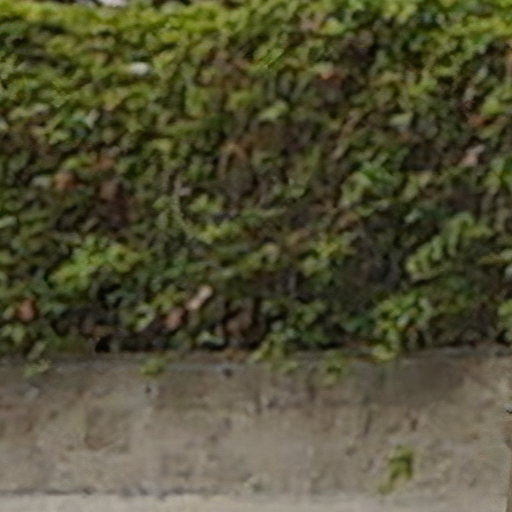}
        \vspace{-0.3cm}
     \end{minipage}
    }%

    \vspace{-3mm}  
    \subfigure[Raw image (visualized)]
    {
     \begin{minipage}{.18\linewidth}
        \centering
        \includegraphics[width=\linewidth]{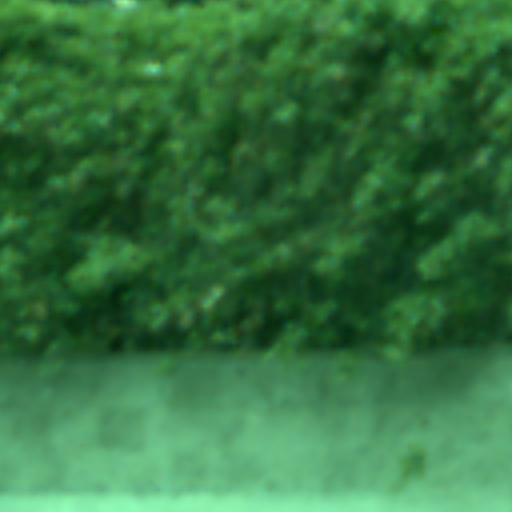}
        \vspace{-0.3cm}
     \end{minipage}
    }%
    \subfigure[Zhang \etal~\cite{SRRAW}]
    {
     \begin{minipage}{.18\linewidth}
        \centering
        \includegraphics[width=\linewidth]{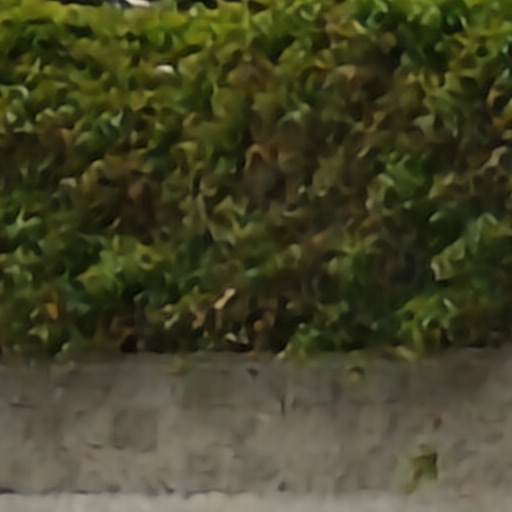}
        \vspace{-0.3cm}
     \end{minipage}
    }%
    \subfigure[Ours]
    {
     \begin{minipage}{.18\linewidth}
        \centering
        \includegraphics[width=\linewidth]{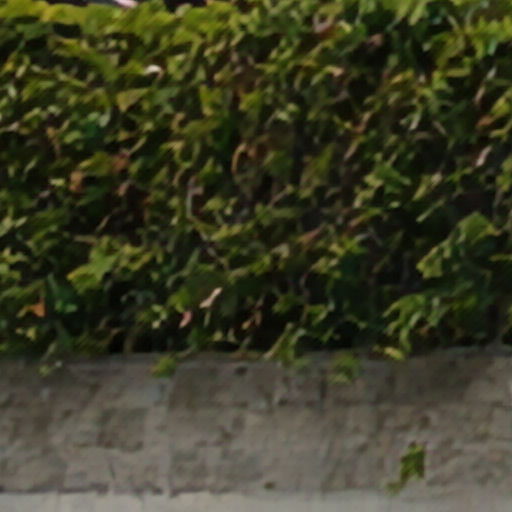}
        \vspace{-0.3cm}
     \end{minipage}
    }%
    \subfigure[Ours (GAN)]
    {
     \begin{minipage}{.18\linewidth}
        \centering
        \includegraphics[width=\linewidth]{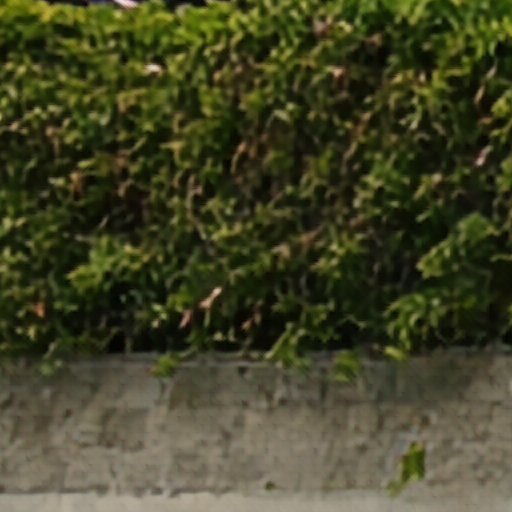}
        \vspace{-0.3cm}
     \end{minipage}
    }%
    \subfigure[GT] 
    {
     \begin{minipage}{.18\linewidth}
        \centering
        \includegraphics[width=\linewidth]{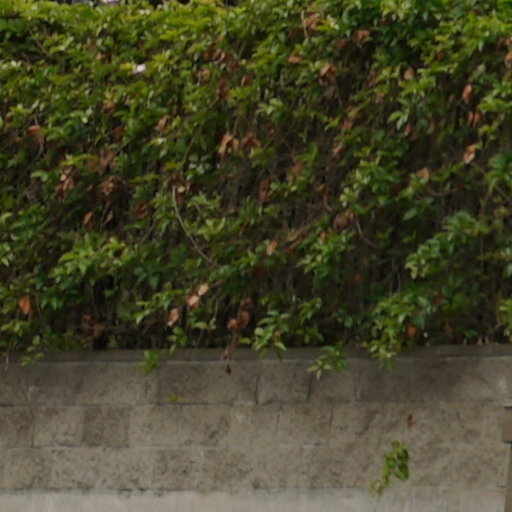}
        \vspace{-0.3cm}
     \end{minipage}
    }%
    \vspace{-4mm}
    \caption{Visual comparison on SR-RAW dataset. $^\dagger$ means that the result is obtained given LR sRGB image as input.}
    \label{fig:SRRAW-result}
    \vspace{-3mm}
  \end{figure*}

  \begin{table*}[t] 
    \small
    \renewcommand\arraystretch{1}
    \caption{Quantitative results on SR-RAW dataset. Methods taking LR sRGB image as input are marked with $^\dagger$.}
    \label{tab:SRRAW-results}
    \vspace{-6mm}
    \begin{center}
      \begin{tabular}{cccc}
        \toprule
        Method & \tabincell{c}{\emph{Original GT} \\ \footnotesize{PSNR}\scriptsize{$\uparrow$} \small{/} \footnotesize{SSIM}\scriptsize{$\uparrow$} \small{/} \footnotesize{LPIPS}\scriptsize{$\downarrow$}}
        & \tabincell{c}{\emph{Align GT with raw} \\ \footnotesize{PSNR}\scriptsize{$\uparrow$} \small{/} \footnotesize{SSIM}\scriptsize{$\uparrow$} \small{/} \footnotesize{LPIPS}\scriptsize{$\downarrow$}}
        & \tabincell{c}{\emph{Align GT with result} \\ \footnotesize{PSNR}\scriptsize{$\uparrow$} \small{/} \footnotesize{SSIM}\scriptsize{$\uparrow$} \small{/} \footnotesize{LPIPS}\scriptsize{$\downarrow$}} \\
          \hline
          SRGAN$^\dagger$~\cite{SRResNet}        & 18.42 / 0.5534 / 0.456    & 19.32 / 0.5999 / 0.419     & 21.89 / 0.6832 / 0.398  \\
          ESRGAN$^\dagger$~\cite{ESRGAN}         & 18.66 / 0.5563 / 0.435   & 19.55 / 0.6018 / 0.411   & 21.99 / 0.6785 / 0.393 \\
          SPSR$^\dagger$~\cite{SPSR}             & 18.64 / 0.5428 / 0.454   & 19.50 / 0.5854 / 0.441   & 21.90 / 0.6603 / 0.425 \\
          RealSR$^\dagger$~\cite{RealSR}         & \textbf{18.69} / 0.5570 / 0.435   & 19.58 / 0.6026 / 0.412   & 22.03 / 0.6796 / 0.394 \\
          Zhang \etal~\cite{SRRAW}     & 16.03 / 0.5184 / 0.517   & 17.43 / 0.5745 / 0.440    & 22.26 / \textbf{0.7205} / 0.372 \\
          \hline
          Ours  & 17.74 / \textbf{0.5572} / 0.430        & 22.00 / \textbf{0.7049} / 0.346          & 22.50 / \textbf{0.7205} / 0.342  \\
          Ours (GAN)    & 17.71 / 0.5501 / \textbf{0.422}        & \textbf{22.10} / 0.6996 / \textbf{0.340}   & \textbf{22.59} / 0.7156 / \textbf{0.336} \\
        \bottomrule
      \end{tabular}
    \end{center}
    \vspace{-8mm}
  \end{table*}

\vspace{-1mm}
\section{Experiments}\label{section:Experiments}
\vspace{-1mm}

\subsection{Implementation Details}
\vspace{-1mm}
  \noindent\textbf{Datasets.}
  We conduct experiments on two datasets, \ie, Zurich RAW to RGB (ZRR)~\cite{PyNet} and SR-RAW~\cite{SRRAW}.

  In the ZRR dataset, 20 thousand image pairs are collected and roughly aligned via SIFT keypoints~\cite{SIFT} and the RANSAC algorithm~\cite{RANSAC}, and the cropped patches with cross-correlation $<0.9$ are discarded, resulting in 48,043 raw-sRGB pairs of size $448 \times 448$.
  We follow the official division to train our LiteISPNet with 46.8k pairs, and report the quantitative results on the remaining 1.2k pairs.

  In the SR-RAW dataset, there are 500 scenes of images collected.
  In each scene, the authors take 7 photos with various focal length (24, 35, 50, 70, 100, 150 and 240 mm), where the 24/100, 35/150 and 50/240 pairs form a 4$\times$ super-resolution dataset (\ie, with 1,500 pairs in total). 
  We use 400 scenes for training, 50 scenes for validation, and report the performance on 35/150 mm pairs of rest 50 scenes.
  For a fair comparison, we replace LiteISPNet by SRResNet~\cite{SRResNet} used in Zhang~\etal~\cite{SRRAW} on the SR-RAW dataset.

  \noindent\textbf{Training Details.}
  During training, data augmentation is applied on training images, including random horizontal flip, vertical flip and $90^\circ$ rotation.
  In each iteration, 16 packed raw patches with 4 channels are extracted as inputs.
  Our framework is optimized by ADAM algorithm~\cite{Adam} with $\beta_1=0.9$ and $\beta_2=0.999$ for 100 epochs.
  The learning rate is initially set to $1\times10^{-4}$ and is decayed to half after 50 epochs.
  The experiments are conducted with PyTorch~\cite{PyTorch} framework on an Nvidia GeForce RTX 2080Ti GPU.

  \noindent\textbf{Evaluation Configurations.}
  To evaluate the performance quantitatively, we compute three metrics on RGB channels, \ie, Peak Signal to Noise Ratio (PSNR), Structural Similarity (SSIM)~\cite{SSIM} and Learned Perceptual Image Patch Similarity (LPIPS)~\cite{LPIPS}.
  Note that in this paper, we use the version 0.1 of LPIPS trained on the AlexNet network.
  All results of the competing methods are generated via the officially released model.
  In addition, we also count the inference time on ZRR dataset to evaluate model efficiency.

 Besides providing the metrics computed with \emph{Original GT}, we additionally provide two sets of metrics for a comprehensive and more accurate comparison by considering the alignment.
 Specifically, we align $\mathbf{y}$ with GCM output $\tilde{\mathbf{y}}$ by PWC-Net~\cite{PWC-Net}, and the metrics computed with such warped $\mathbf{y}$ are denoted by \emph{Align GT with raw}.
 In addition, considering that previous models trained with misaligned data may cause pixel shift in the result, we further align the ground-truth with the output of each method, and the metrics are denoted by \emph{Align GT with result}.

    \vspace{-1mm}  
\subsection{Experimental Results on ZRR Dataset}
    \vspace{-1mm}  
  To evaluate the effectiveness of the proposed method on ZRR dataset, we compare our model with three state-of-the-art methods, \ie, PyNet~\cite{PyNet}, AWNet~\cite{AWNet} and MW-ISPNet~\cite{AIM2020ISP}.
  Note that AWNet (raw) and AWNet (demosaicked) denote two models proposed in AWNet, which take 4-channel raw image and 3-channel demosaicked image as input respectively, and MW-ISPNet (GAN) denotes MW-ISPNet trained with adversarial loss.

  As shown in Table~\ref{tab:ZRR-results}, LiteISPNet exceeds the competing methods on all metrics in \emph{Align GT with raw} and \emph{Align GT with result}.
  Furthermore, when training with adversarial loss, our LiteISPGAN achieves better LPIPS score (as well as the PSNR and SSIM indices) than MW-ISPNet (GAN), which is the winner of perceptual track in the AIM 2020 Learned Smartphone ISP challenge~\cite{AIM2020ISP}.
  It is worth mentioning that our model achieves the superior performance with a lightweight structure (the number of parameters is $\sim$25\% and $\sim$40\% of AWNet and MW-ISPNet), and the inference time is only $\sim$23 ms for a $448 \times 448$ input ($\sim$40\% and $\sim$20\% of AWNet and MW-ISPNet).

  Besides, we show the qualitative results in Fig.~\ref{fig:ZRR-results1}.
  It can be seen that the results of PyNet, AWNet (demosaicked) and MW-ISPNet (GAN) are blurry.
  AWNet (raw) is able to retain more details, however, it may generate artifacts and the result is less satisfactory.
  In contrast, our results are visually more pleasant while preserving finer details.
  Please refer to the suppl. for more results.

    \vspace{-1mm}  
  \subsection{Experimental Results on SR-RAW Dataset}
    \vspace{-1mm}  
  %
  %
  The proposed method is also evaluated on SR-RAW dataset for $4\times$ SR.
  Apart from the raw image SR method (Zhang~\etal~\cite{SRRAW}), we also compare with state-of-the-art sRGB image SR methods (\ie, SRGAN~\cite{SRResNet}, ESRGAN~\cite{ESRGAN}, SPSR~\cite{SPSR} and RealSR~\cite{RealSR}), where low-resolution sRGB images provided in SR-RAW dataset are taken as input.
  The quantitative and qualitative results are given in Table~\ref{tab:SRRAW-results} and Fig.~\ref{fig:SRRAW-result}, respectively.

  We can see that SRGAN~\cite{SRResNet}, ESRGAN~\cite{ESRGAN}, SPSR~\cite{SPSR} and RealSR~\cite{RealSR} tend to generate noisy results with undesired textures, and show unsatisfactory quantitative performance.
  Although trained with mild misalignment robust contextual bilateral (CoBi) loss, Zhang~\etal~\cite{SRRAW} is unable to recover fine details and resulting in blurry results.
  Our method, with the same SRResNet~\cite{SRResNet} backbone as Zhang~\etal~\cite{SRRAW}, can preserve more textures \ and \ generate
  \clearpage

\begin{figure*}[t]
	\vspace{-6mm}
	\centering
	\subfigure[Full raw image (visualized)]
	{
		\begin{minipage}{0.2\linewidth}
			\centering
			\includegraphics[width=\linewidth]{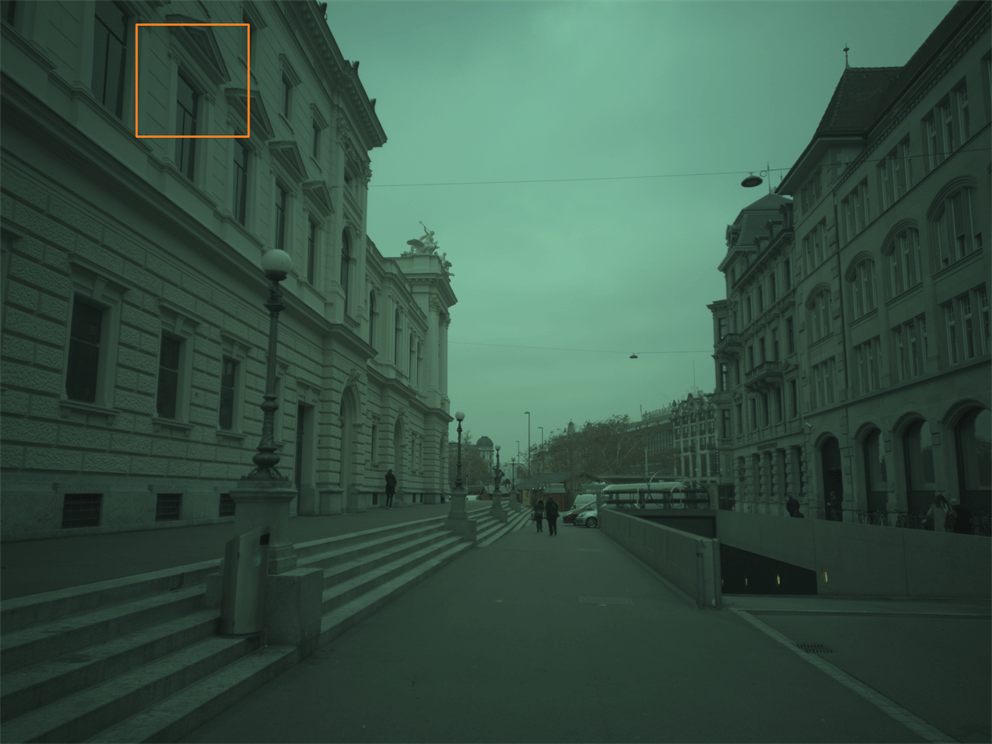}
			\vspace{-0.3cm}
		\end{minipage}
	}%
	\subfigure[SPN] 
	{
		\begin{minipage}{0.15\linewidth}
			\centering
			\includegraphics[width=\linewidth]{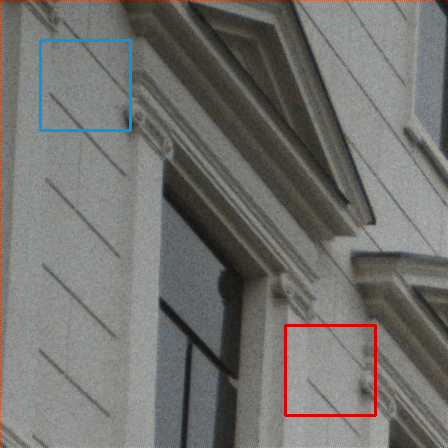}
			\vspace{-0.3cm}
		\end{minipage}
	}%
	\subfigure[SPN+$\mathbf{y}$] 
	{
		\begin{minipage}{0.15\linewidth}
			\centering
			\includegraphics[width=\linewidth]{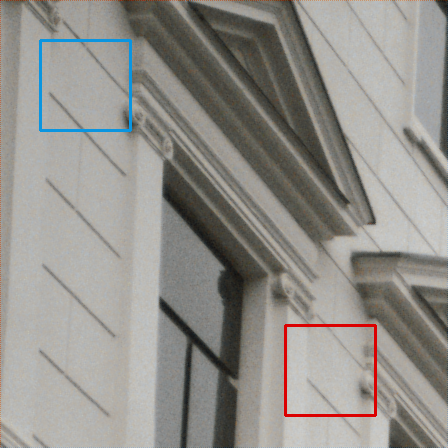}
			\vspace{-0.3cm}
		\end{minipage}
	}%
	\subfigure[SPN+$\mathbf{y}$+$\bm{\tau}$(Ours)] 
	{
		\begin{minipage}{0.15\linewidth}
			\centering
			\includegraphics[width=\linewidth]{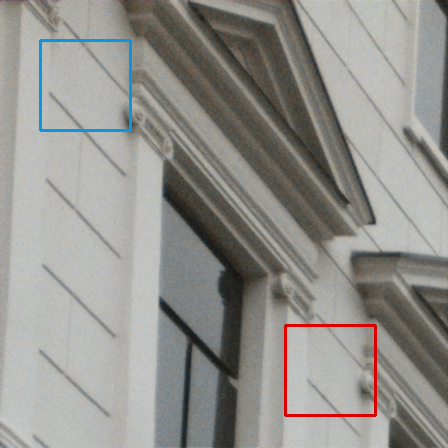}
			\vspace{-0.3cm}
		\end{minipage}
	}%
	\subfigure[GT] 
	{
		\begin{minipage}{0.15\linewidth}
			\centering
			\includegraphics[width=\linewidth]{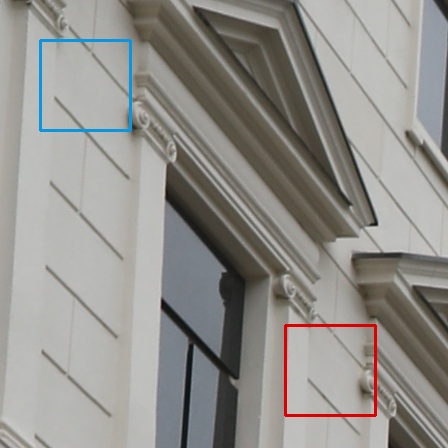}
			\vspace{-0.3cm}
		\end{minipage}
	}%
  \vspace{-4mm}
	\caption{Visual results of GCM output ($\tilde{\mathbf{y}}$). With the guidance of $\mathbf{y}$, the color of (c)(d) is closer to GT. Dark corner can also be observed, as the patch in blue box is darker in (b)(c) but has similar illumination with the patch in red box in (d)(e).}
	\label{fig:ZRR-ablation-GCM}
  \vspace{-6mm}
\end{figure*}

\stripsep 0pt
\begin{strip}
	\begin{minipage}[h]{\linewidth}
		\small
		\begin{minipage}[t]{0.295\linewidth}
			\centering
			\captionsetup{font={small}}
			\captionof{table}{\centering Ablation study on alignment strategies.}
			\label{tab:ZRR-ablation1}
			\vspace{-3mm}
			\begin{tabular}{@{\ }l@{\ \ }c@{\ \ }c@{\ }}
				\toprule
				Method & PSNR & LPIPS \\
				\midrule
				SIFT (baseline) & 23.49 & 0.148 \\ 
				Align $\mathbf{y}$ with $\hat{\mathbf{y}}$ & 23.33 & 0.136 \\ 
				Align $\mathbf{y}$ with $\hat{\mathbf{x}}$ & 23.52 & 0.135 \\ 
				Align $\mathbf{y}$ with $\tilde{\mathbf{y}}$ (Ours) & 23.76 & 0.133 \\  
				\bottomrule
			\end{tabular}
		\end{minipage}
		\hspace{2mm}
		\begin{minipage}[t]{0.395\linewidth}
			\centering
			\captionsetup{font={small}}
			\captionof{table}{Ablation study on GCM. PSNR is calculated with GCM output ($\tilde{\mathbf{y}}$) and final result ($\hat{\mathbf{y}}$), respectively.}
			\label{tab:ZRR-ablation2}
			\vspace{-3mm}
			\begin{tabular}{@{\ }l@{\ \ \ }c@{\ \ \ }c@{\ }}
				\toprule
				GCM Components & PSNR ($\tilde{\mathbf{y}}$) & PSNR ($\hat{\mathbf{y}}$) \\
				\midrule
				N/A & - & 23.49\\
				SPN & 20.67 & 23.61 \\
				SPN + $\mathbf{y}$ & 26.33 & 23.69 \\
				SPN + $\mathbf{y}$ + $\bm{\tau}$ (Ours) & 26.93 & 23.76\\
				\bottomrule
			\end{tabular}
		\end{minipage}
		\hspace{2mm}
		\begin{minipage}[t]{0.26\linewidth}
			\centering
			\captionsetup{font={small}}
			\captionof{table}{\centering Ablation study on LiteISPNet.}
			\label{tab:ZRR-ablation2}
      \vspace{-3mm}
      \begin{tabular}{@{\ \ }c@{\ \ \ }c@{\ \ \ }c@{\ }}
				\toprule
				\multirow{1}{*}{\# RCAB} & \multirow{1}{*}{\tabincell{c}{Time (ms)}} & \multirow{1}{*}{PSNR} \\
				\midrule
				2 & 15.3 & 23.54 \\
				4 & 23.3 & 23.76 \\
        8 & 38.0 & 23.74 \\
        20 & 91.5 & 23.79 \\
				\bottomrule
			\end{tabular}
		\end{minipage}
	\end{minipage}
	\vspace{2mm}
\end{strip}

\noindent neat results. More qualitative results are given in the suppl.

    \vspace{-2mm}  
\section{Ablation Study}
    \vspace{-2mm}  
  In this section, we conduct extensive ablation studies on the proposed joint learning framework, and report the PSNR metric of \emph{Align GT with raw} on the ZRR dataset.

    \vspace{-1.5mm}  
\subsection{Alignment Strategies}
    \vspace{-1.3mm}  
  To solve the misalignment issue between input raw and target sRGB image, an intuitive way is to align them.
  For example, image pairs in ZRR dataset are roughly aligned by SIFT~\cite{SIFT} algorithm.
  We train a LiteISPNet with such image pairs as a baseline (see Table~\ref{tab:ZRR-ablation1}), and evaluate several potential strategies for better alignment.
  (1)~Learning image alignment jointly with RAW-to-sRGB mapping may lead to an iterative optimization process, so we align $\mathbf{y}$ with the output of LiteISPNet ($\hat{\mathbf{y}}$) during training.
  Unfortunately, RAW-to-sRGB mapping is not a pixel-wise operation, resulting in more freedom for optical flow estimation and may cause more severe pixel shift.
  (2)~We also estimate an optical flow between the demosaicked image $\hat{\mathbf{x}}$ and $\mathbf{y}$, and align $\mathbf{y}$ with $\hat{\mathbf{x}}$. Pixel shift is alleviated to some extent, yet the quality is still limited due to the color inconsistency.
  (3)~With proposed GCM module, we can obtain $\tilde{\mathbf{y}}$ whose color is consistent with $\mathbf{y}$ while the pixel positions are same with $\hat{\mathbf{x}}$. Thus, our method performs favorably against other alignment strategies.
  Please refer to the suppl. for visual results.

    \vspace{-1.5mm}  
\subsection{GCM Module}
    \vspace{-1.2mm}  
  
  To evaluate each individual component of GCM module, we further perform experiments as shown in Table~\ref{tab:ZRR-ablation2}.
  The baseline is a LiteISPNet trained with mildly aligned image pairs in the dataset.
  When adding an SPN to the baseline, a 0.12 dB PSNR gain is attained owing to a rough color correction.
  By further introducing the guidance provided by target sRGB image $\mathbf{y}$ and concatenating the coordinate map $\bm{\tau}$, the quality of GCM output $\tilde{\mathbf{y}}$ is effectively improved (see Fig.~\ref{fig:ZRR-ablation-GCM}), which helps in optical flow estimation and providing better aligned supervision.
  As a result, the performance of the LiteISPNet is also promoted.
  Note that the ground-truth $\mathbf{y}$ is applied to generate the GCM output, leading to a higher PSNR metric than the LiteISPNet output. However, it is unavailable during inference of the LiteISPNet.

    \vspace{-1.5mm}  
  \subsection{Structure of LiteISPNet}
    \vspace{-1mm}  
  To explore the structure of LiteISPNet, we also conducted experiments on the number of residual channel attention blocks (RCABs) in each residual group (RG).
  As shown in Table~\ref{tab:ZRR-ablation2}, using 4 RCABs is sufficient with the well aligned training data, and deeper networks do not bring noticeable performance improvements. 
  Therefore, we apply 4 RCABs to achieve a better efficiency-performance tradeoff in our LiteISPNet.

\vspace{-2mm}  
\section{Conclusion}\label{section:Conclusion}
\vspace{-2mm}  
Learning with inaccurately aligned supervision is prone to causing pixel shift and generating blurry results, but existing methods usually fail to solve the inherent misalignment problem in many RAW-to-sRGB tasks due to the severe color inconsistency between raw and sRGB pairs.
To diminish the effect of color inconsistency, we presented a global color mapping (GCM) module, where SPN is leveraged to avoid spatial shift of pixels, and the target sRGB image serves as a guidance to convert the color of the raw data.
Then, a pre-trained optical flow estimation model (\eg, PWC-Net) is deployed for obtaining well aligned supervision, which is used to train the RAW-to-sRGB mapping in a joint learning manner.
Extensive experiments on ZRR and SR-RAW datasets show that our proposed method can achieve better performance against state-of-the-art methods both quantitatively and qualitatively.

\vspace{-2mm}
\section*{Acknowledgement}
\vspace{-2mm}

This work was supported by the National Natural Science Foundation of China (NSFC) under Grants No.s U19A2073 and 62006064.

{\small
\bibliographystyle{ieee_fullname}
\bibliography{egbib}
}

\clearpage

\renewcommand{\thesection}{\Alph{section}}
\renewcommand{\thetable}{\Alph{table}}
\renewcommand{\thefigure}{\Alph{figure}}

\setcounter{section}{0}
\setcounter{table}{0}
\setcounter{figure}{0}
%
\section{Content}
The content of this supplementary material involves:
\begin{itemize}
    \vspace{-.5em}
    \item{Network structure of GCM, LiteISPNet and the discriminator of LiteISPGAN in Sec.~\ref{sec:NetworkStructure}.}
    \vspace{-.5em}
    \item{Visual results of alignment in Sec.~\ref{sec:alignment}.}
    \vspace{-.5em}
    \item{Qualitative results of ablation study in Sec.~\ref{sec:ablation}}
    \vspace{-.5em}
    \item{Implementation details on SR-RAW dataset in Sec.~\ref{sec:SRRAW}.}
    \vspace{-.5em}
    \item{Quantitative results for re-splitting the train/test set on ZRR dataset in Sec.~\ref{sec:ZRR}.}
    \vspace{-.5em}
    \item{Additional visual comparison results on SR-RAW and ZRR dataset in Sec.~\ref{sec:Vis}.}
    \vspace{-.5em}
\end{itemize}
%
\section{Network Structure}\label{sec:NetworkStructure}
    %
	Global color mapping (GCM) module involves two components: spatially preserving network (SPN) and GuideNet.
	SPN stacks $1\times1$ convolutional layers to guarantee spatial independence of the mapping and GuideNet generates a conditional guidance vector from the target sRGB to modulate SPN features.
	The detailed structure of GCM are shown in Table~\ref{tab:GCM_model}.
	
	The structure configuration of LiteISPNet are shown in Table~\ref{tab:LiteISPNet}.
	LiteISPNet is a U-Net~\cite{U-Net} based multi-level wavelet ISP network.
  In each residual group, we only apply 4 residual channel attention blocks (RCABs).
    
  The discriminator structure of LiteISPGAN are shown in Table~\ref{tab:Discriminator}.
  We apply 54 × 54 PatchGAN~\cite{CycleGAN}, which distinguishes whether the image patch is real or fake. 

\section{Visual Results of Alignment}\label{sec:alignment}
\begin{figure*}[t] 
	\centering
	\subfigure[Demosaicked raw ($\hat{\mathbf{x}}$)]
	{
		\begin{minipage}{.19\linewidth}
			\centering
			\includegraphics[width=\linewidth]{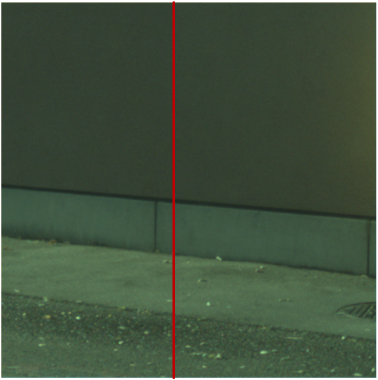}
			
			\includegraphics[width=\linewidth]{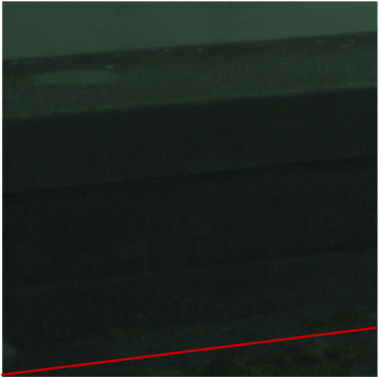}
			
			\includegraphics[width=\linewidth]{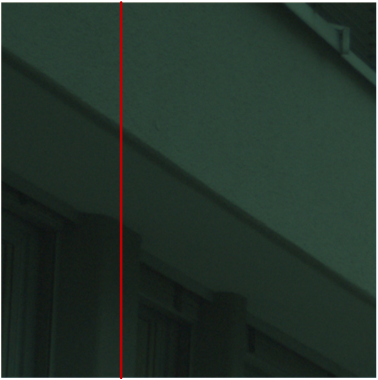}
			\vspace{-0.2cm}
		\end{minipage}
	}%
	\subfigure[GCM output ($\tilde{\mathbf{y}}$)]
	{
		\begin{minipage}{.19\linewidth}
			\centering
			\includegraphics[width=\linewidth]{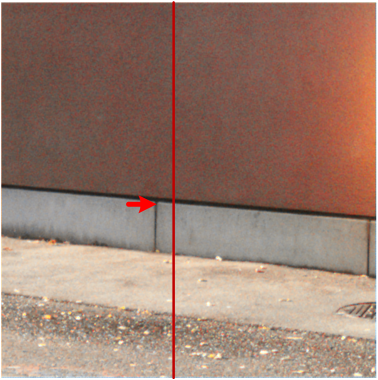}
			
			\includegraphics[width=\linewidth]{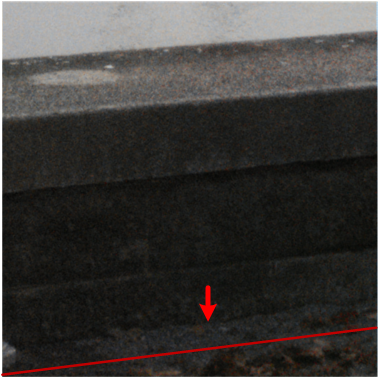}
			
			\includegraphics[width=\linewidth]{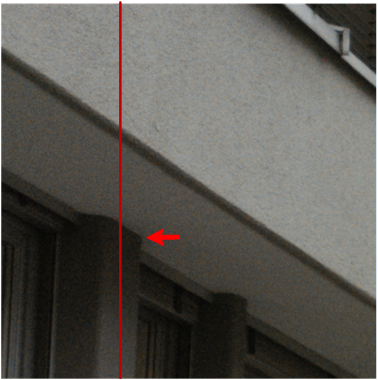}
			\vspace{-0.2cm}
		\end{minipage}
	}%
	\subfigure[LiteISPNet output ($\hat{\mathbf{y}}$)]
	{
		\begin{minipage}{.19\linewidth}
			\centering
		    \includegraphics[width=\linewidth]{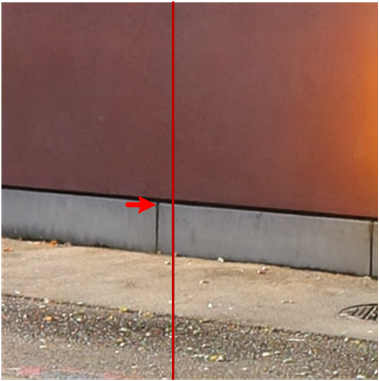}
			
			\includegraphics[width=\linewidth]{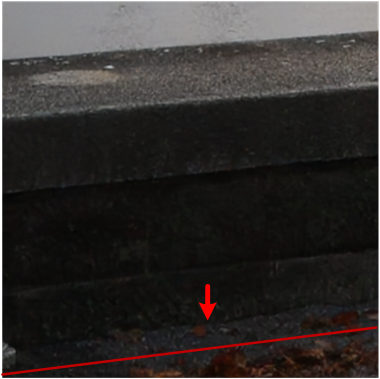}
			
			\includegraphics[width=\linewidth]{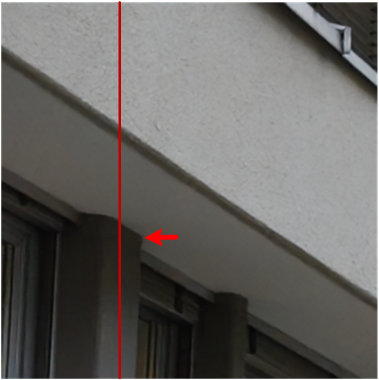}
			\vspace{-0.2cm}
		\end{minipage}
	}%
	\subfigure[Warped target sRGB ($\mathbf{y}^\mathit{w}$)]
	{
		\begin{minipage}{.19\linewidth}
			\centering
			\includegraphics[width=\linewidth]{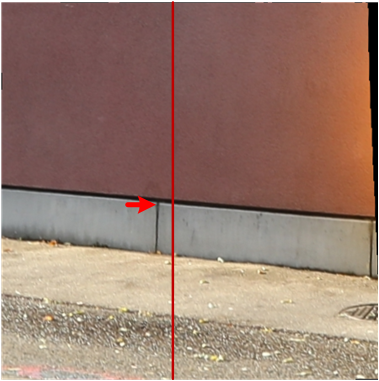}
			
			\includegraphics[width=\linewidth]{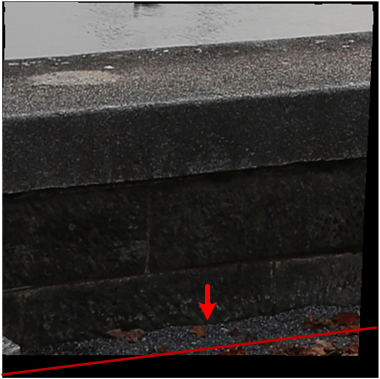}
			
			\includegraphics[width=\linewidth]{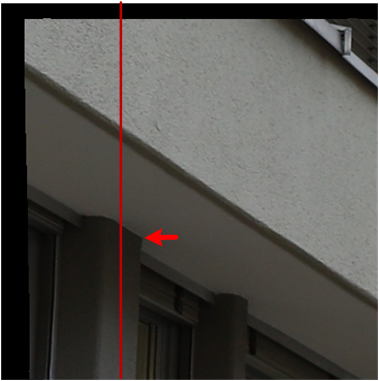}
			\vspace{-0.2cm}
		\end{minipage}
	}%
	\subfigure[GT ($\mathbf{y}$)]
	{
		\begin{minipage}{.19\linewidth}
			\centering
			\includegraphics[width=\linewidth]{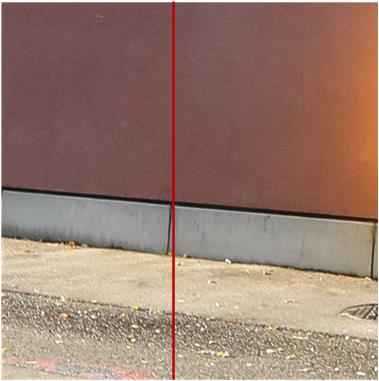}
			
			\includegraphics[width=\linewidth]{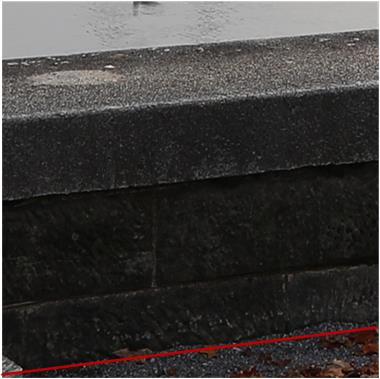}
			
			\includegraphics[width=\linewidth]{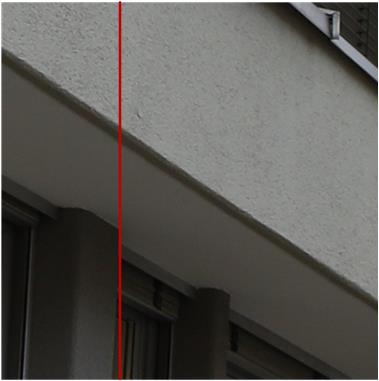}
			\vspace{-0.2cm}
		\end{minipage}
	}%
  \caption{Alignment visual results obtained by our joint learning framework.
  With the reference line, it can be observed our method obtains the well aligned data pairs 
  while the demosaicked raw is not aligned with GT.}
	\label{fig:ZRR-listresults}
\end{figure*}

We show the demosaicked raw image ($\hat{\mathbf{x}}$), GCM output ($\tilde{\mathbf{y}}$), LiteISPNet output ($\hat{\mathbf{y}}$), warped target sRGB image ($\mathbf{y}^\mathit{w}$) and the original target sRGB image ($\mathbf{y}$) in Fig.~\ref{fig:ZRR-listresults}. 

It can be seen that the color of $\tilde{\mathbf{y}}$ is consistent with $\mathbf{y}$. 
Although GCM model cannot perform local operations (\eg, denoising), benefiting from PWC-Net~\cite{PWC-Net}, we can still align $\mathbf{y}$ with $\tilde{\mathbf{y}}$ robustly.
Under the supervision of well aligned training data, LiteISPNet output has almost no pixel shift.
In short, our method achieves the joint of image alignment and RAW-to-sRGB mapping.

\section{Qualitative Results on Ablation Study}\label{sec:ablation}
We show more qualitative results of different alignment strategies in Fig.~\ref{fig:ZRR-ablation-alignment}. 

Due to the limitation of space in the submitted manuscript, we only showed the outputs of GCM model in Sec. 5.2.
Here, by visualizing the illuminance ratio between the output of GCM and ground-truth (GT), we show the influence of each component and the dark corner phenomenon more clearly in Fig.~\ref{fig:ZRR-ablation-GCM}.

\section{Implementation Details on SR-RAW Dataset}\label{sec:SRRAW}
%

In each image pair of the SR-RAW dataset, the short focal-length raw image is used as input, while the long focal-length sRGB image is adopted as the ground-truth.
In order to align high-resolution (HR) sRGB images with low-resolution (LR) raw images, we adopt downsampled HR sRGB image $\mathbf{y}_\downarrow$ to generate the conditional guidance vector in the GCM model. 
Then the optical flow between the GCM output $\tilde{\mathbf{y}}$ and $\mathbf{y}_\downarrow$ is estimated. Note that the size of optical flow is a quarter of the HR sRGB image. 
Thus, we upsample the optical flow to get the warped HR sRGB image. 
Finally, the warped HR sRGB image is utilized to supervise the learning of the backbone (SRResNet~\cite{SRResNet}).

Following~\cite{SRRAW}, we use 400 scenes of images for training, 50 for validation, and the rest 50 for testing, and report the performance on 35/150 mm pairs in Table~{\color{red}2} of the main text.
For a comprehensive comparison, we further show the performance on all 24/100, 35/150 and 50/240 test pairs in Table~\ref{tab:SRRAW}. 
It can be seen that our method can still achieve better quantitative performance against all competing methods.

\section{Quantitative Results for Re-splitting the Train/Test Set on ZRR Dataset}\label{sec:ZRR}
For the ZRR dataset, we follow the official division to train our LiteISPNet with 46,839 pairs, and report the quantitative results on the remaining 1,204 pairs in the main text.
Here we conducted an experiment by re-splitting the dataset at approximately $9:1$, \ie, 43,200 pairs for training and the rest 4,843 pairs for testing.
Table~\ref{tab:ZRR} shows the quantitative results, and it can be seen that our LiteISPNet also exceeds the competing methods.

\section{Additional Visual Comparison Results on SR-RAW and ZRR Dataset}\label{sec:Vis}
\begin{figure*}[t] 
     \centering
     \subfigure[Bicubic$^\dagger$]
     {
      \begin{minipage}{.19\linewidth}
         \centering
         \includegraphics[width=\linewidth]{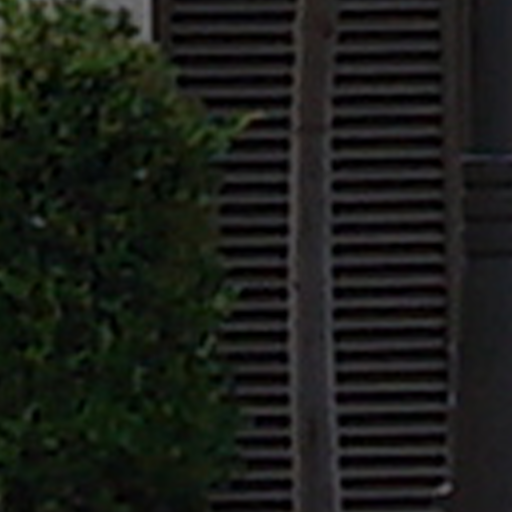}
         \vspace{-0.3cm}
      \end{minipage}
     }%
     \subfigure[SRGAN$^\dagger$~\cite{SRResNet}]
     {
      \begin{minipage}{.19\linewidth}
         \centering
         \includegraphics[width=\linewidth]{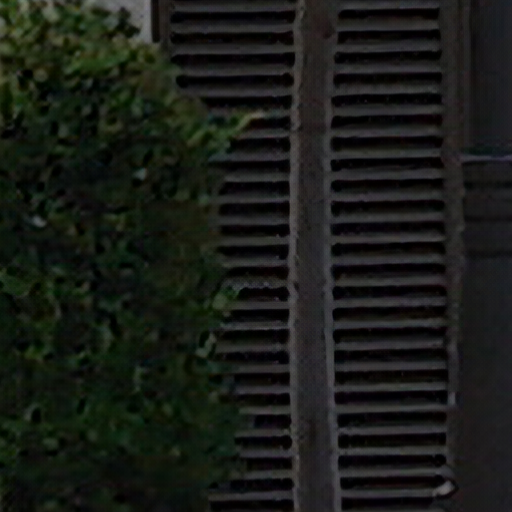}
         \vspace{-0.3cm}
      \end{minipage}
     }%
     \subfigure[ESRGAN$^\dagger$~\cite{ESRGAN}]
     {
      \begin{minipage}{.19\linewidth}
         \centering
         \includegraphics[width=\linewidth]{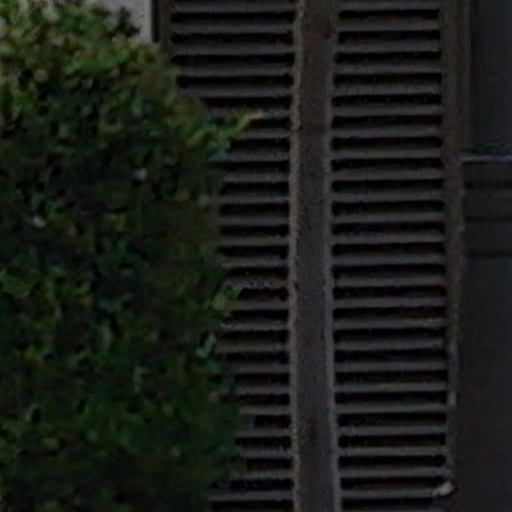}
         \vspace{-0.3cm}
      \end{minipage}
     }%
     \subfigure[SPSR$^\dagger$~\cite{SPSR}]
     {
      \begin{minipage}{.19\linewidth}
         \centering
         \includegraphics[width=\linewidth]{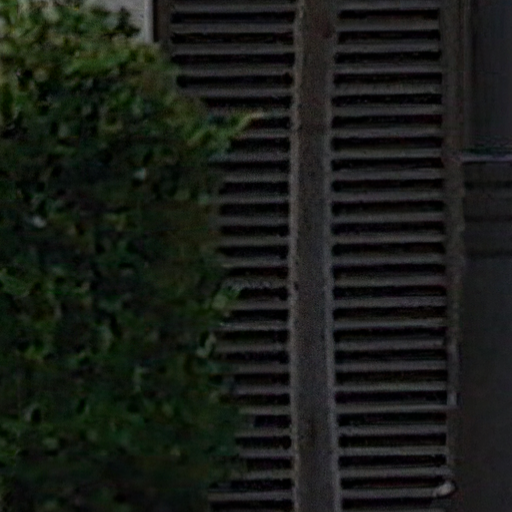}
         \vspace{-0.3cm}
      \end{minipage}
     }%
     \subfigure[RealSR$^\dagger$~\cite{RealSR}]
     {
      \begin{minipage}{.19\linewidth}
         \centering
         \includegraphics[width=\linewidth]{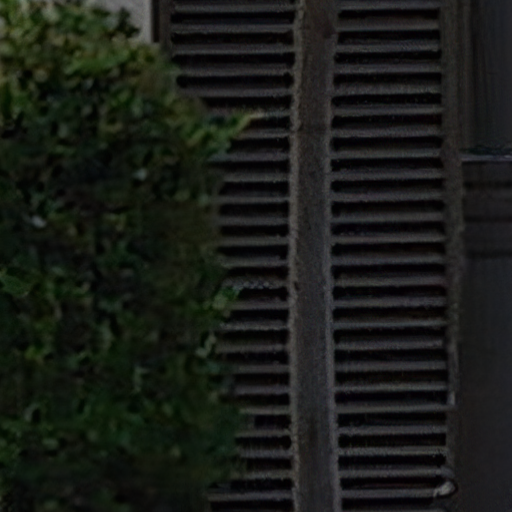}
         \vspace{-0.3cm}
      \end{minipage}
     }%

     \subfigure[Raw image (visualized)]
     {
      \begin{minipage}{.19\linewidth}
         \centering
         \includegraphics[width=\linewidth]{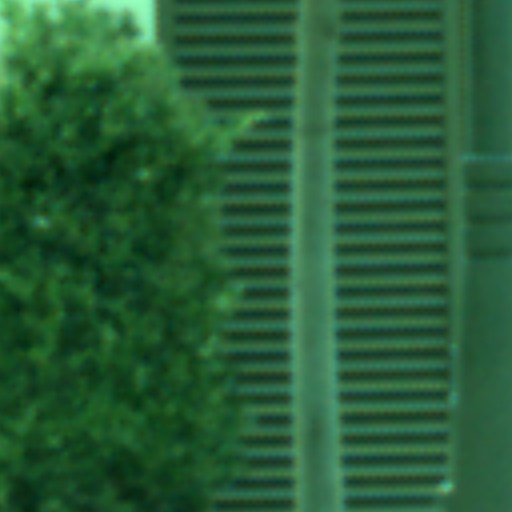}
         \vspace{-0.3cm}
      \end{minipage}
     }%
     \subfigure[Zhang \etal~\cite{SRRAW}]
     {
      \begin{minipage}{.19\linewidth}
         \centering
         \includegraphics[width=\linewidth]{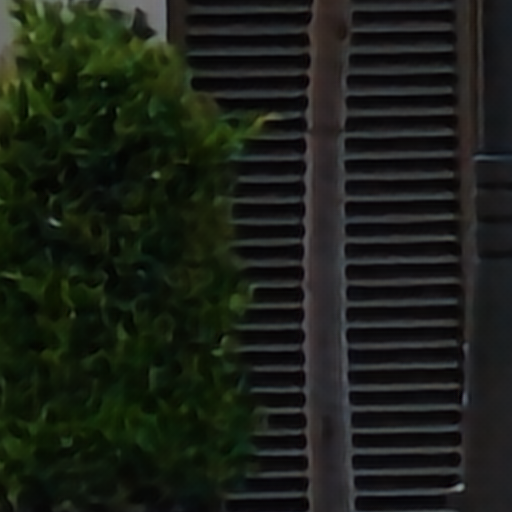}
         \vspace{-0.3cm}
      \end{minipage}
     }%
     \subfigure[Ours]
     {
      \begin{minipage}{.19\linewidth}
         \centering
         \includegraphics[width=\linewidth]{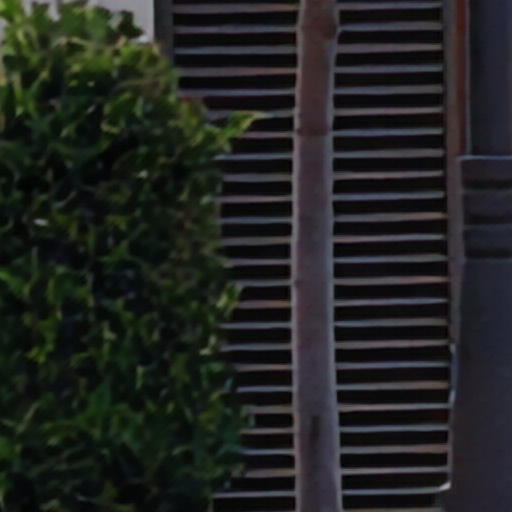}
         \vspace{-0.3cm}
      \end{minipage}
     }%
     \subfigure[Ours (GAN)]
     {
      \begin{minipage}{.19\linewidth}
         \centering
         \includegraphics[width=\linewidth]{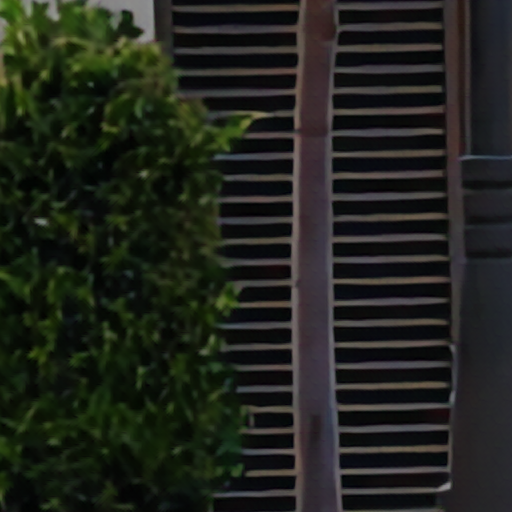}
         \vspace{-0.3cm}
      \end{minipage}
     }%
     \subfigure[GT] 
     {
      \begin{minipage}{.19\linewidth}
         \centering
         \includegraphics[width=\linewidth]{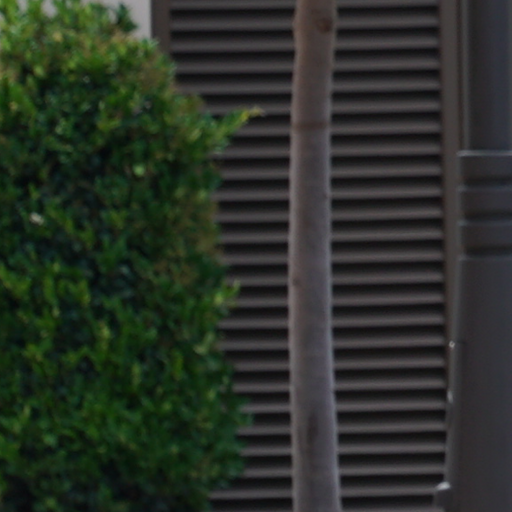}
         \vspace{-0.3cm}
      \end{minipage}
     }%
     \caption{Visual comparison on SR-RAW dataset. $^\dagger$ means that the result is obtained given LR sRGB image as input.
     Our results have more textures on the leaves.}
     \label{fig:SRRAW-result1}
   \end{figure*}

\begin{figure*}[t] 
     \centering
     \subfigure[Bicubic$^\dagger$]
     {
      \begin{minipage}{.19\linewidth}
         \centering
         \includegraphics[width=\linewidth]{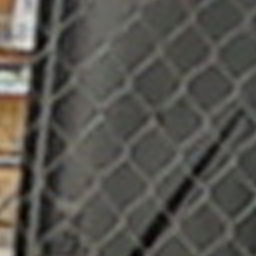}
         \vspace{-0.3cm}
      \end{minipage}
     }%
     \subfigure[SRGAN$^\dagger$~\cite{SRResNet}]
     {
      \begin{minipage}{.19\linewidth}
         \centering
         \includegraphics[width=\linewidth]{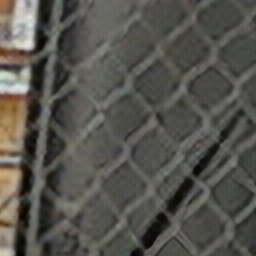}
         \vspace{-0.3cm}
      \end{minipage}
     }%
     \subfigure[ESRGAN$^\dagger$~\cite{ESRGAN}]
     {
      \begin{minipage}{.19\linewidth}
         \centering
         \includegraphics[width=\linewidth]{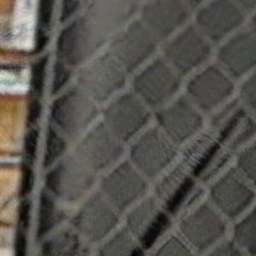}
         \vspace{-0.3cm}
      \end{minipage}
     }%
     \subfigure[SPSR$^\dagger$~\cite{SPSR}]
     {
      \begin{minipage}{.19\linewidth}
         \centering
         \includegraphics[width=\linewidth]{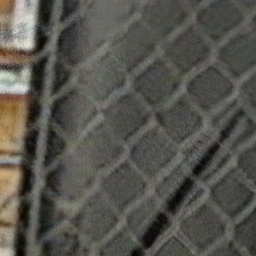}
         \vspace{-0.3cm}
      \end{minipage}
     }%
     \subfigure[RealSR$^\dagger$~\cite{RealSR}]
     {
      \begin{minipage}{.19\linewidth}
         \centering
         \includegraphics[width=\linewidth]{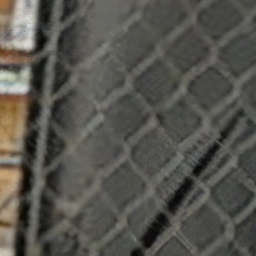}
         \vspace{-0.3cm}
      \end{minipage}
     }%

     \subfigure[Raw image (visualized)]
     {
      \begin{minipage}{.19\linewidth}
         \centering
         \includegraphics[width=\linewidth]{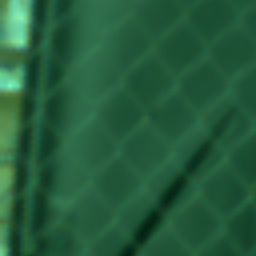}
         \vspace{-0.3cm}
      \end{minipage}
     }%
     \subfigure[Zhang \etal~\cite{SRRAW}]
     {
      \begin{minipage}{.19\linewidth}
         \centering
         \includegraphics[width=\linewidth]{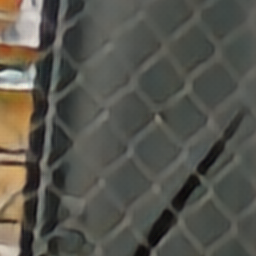}
         \vspace{-0.3cm}
      \end{minipage}
     }%
     \subfigure[Ours]
     {
      \begin{minipage}{.19\linewidth}
         \centering
         \includegraphics[width=\linewidth]{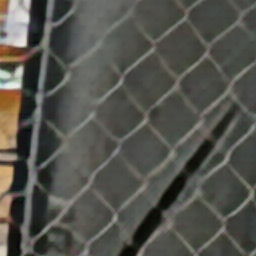}
         \vspace{-0.3cm}
      \end{minipage}
     }%
     \subfigure[Ours (GAN)]
     {
      \begin{minipage}{.19\linewidth}
         \centering
         \includegraphics[width=\linewidth]{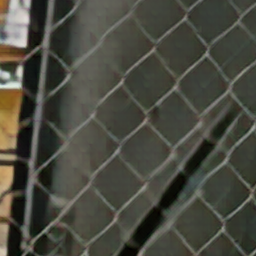}
         \vspace{-0.3cm}
      \end{minipage}
     }%
     \subfigure[GT] 
     {
      \begin{minipage}{.19\linewidth}
         \centering
         \includegraphics[width=\linewidth]{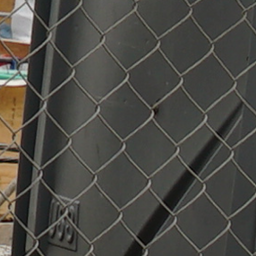}
         \vspace{-0.3cm}
      \end{minipage}
     }%
     \caption{Visual comparison on SR-RAW dataset. $^\dagger$ means that the result is obtained given LR sRGB image as input.
     The edges of our results are sharper.}
     \label{fig:SRRAW-result6}
  \end{figure*}

In Fig.~\ref{fig:SRRAW-result1}$\sim$~\ref{fig:SRRAW-result6}, we show more qualitative comparison results generated by SRGAN~\cite{SRResNet}, ESRGAN~\cite{ESRGAN}, SPSR~\cite{SPSR}, RealSR~\cite{RealSR}, Zhang \etal~\cite{SRRAW} and our models on the SR-RAW dataset. 

\begin{figure*}[t] 
    \centering
    \subfigure[Raw image (visualized)]
    {
      \begin{minipage}{.25\linewidth}
        \centering
        \includegraphics[width=\linewidth]{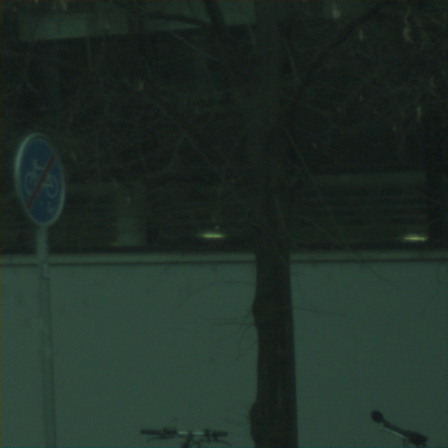}
        \vspace{-0.3cm}
      \end{minipage}
    }%
    \subfigure[PyNet~\cite{PyNet}]
    {
      \begin{minipage}{.25\linewidth}
        \centering
        \includegraphics[width=\linewidth]{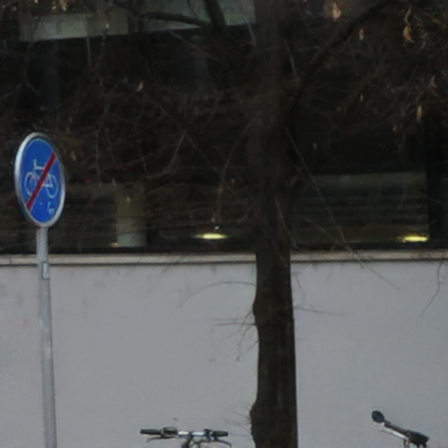}
        \vspace{-0.3cm}
      \end{minipage}
    }%
    \subfigure[AWNet (raw)~\cite{AWNet}]
    {
      \begin{minipage}{.25\linewidth}
        \centering
        \includegraphics[width=\linewidth]{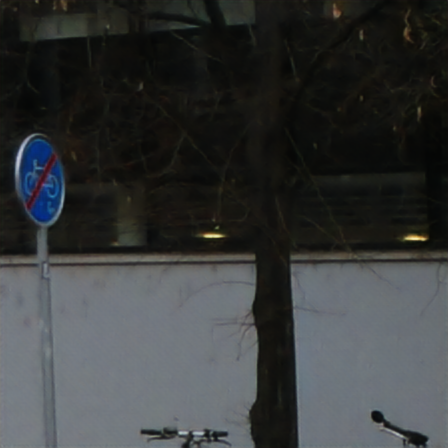}
        \vspace{-0.3cm}
      \end{minipage}
    }%
    \subfigure[AWNet (demosaicked)~\cite{AWNet}]
    {
      \begin{minipage}{.25\linewidth}
        \centering
        \includegraphics[width=\linewidth]{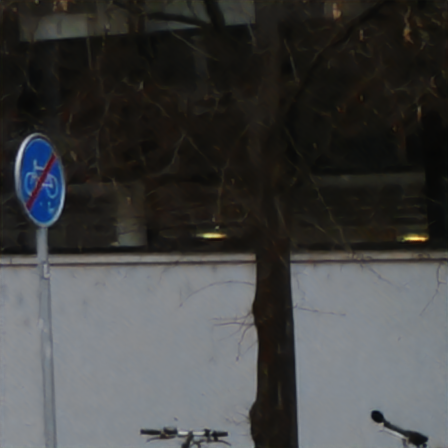}
        \vspace{-0.3cm}
      \end{minipage}
    }%
    
    \subfigure[MW-ISPNet (GAN)~\cite{AIM2020ISP}]
    {
      \begin{minipage}{.25\linewidth}
        \centering
        \includegraphics[width=\linewidth]{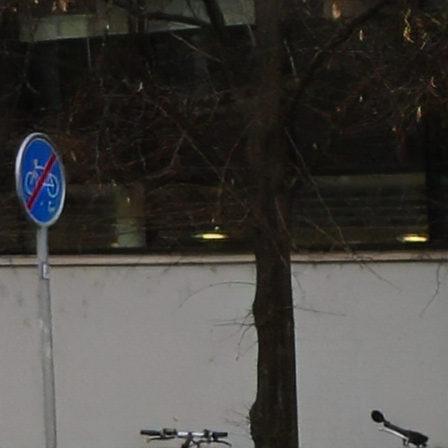}
        \vspace{-0.3cm}
      \end{minipage}
    }%
    \subfigure[Ours (LiteISPNet)]
    {
      \begin{minipage}{.25\linewidth}
        \centering
        \includegraphics[width=\linewidth]{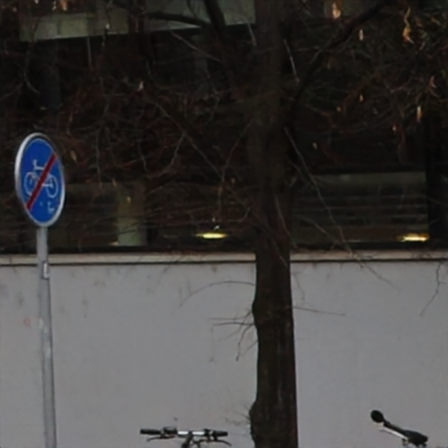}
        \vspace{-0.3cm}
      \end{minipage}
    }%
    \subfigure[Ours (LiteISPGAN)]
    {
      \begin{minipage}{.25\linewidth}
        \centering
        \includegraphics[width=\linewidth]{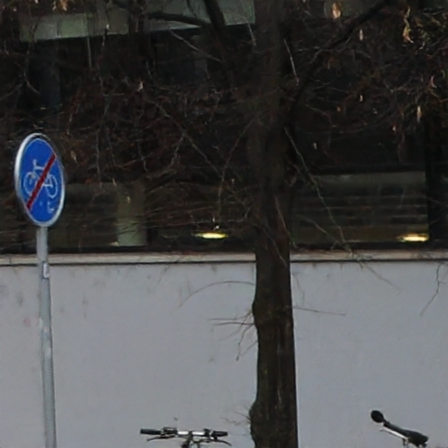}
        \vspace{-0.3cm}
      \end{minipage}
    }%
    \subfigure[GT]
    {
      \begin{minipage}{.25\linewidth}
        \centering
        \includegraphics[width=\linewidth]{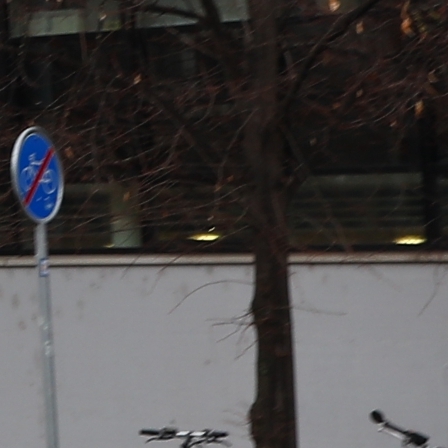}
        \vspace{-0.3cm}
      \end{minipage}
    }%
    \caption{Visual comparisons on ZRR dataset. The tree branches in our results are clearer.}
    \label{fig:ZRR-results4}
    \end{figure*}

In Fig.~\ref{fig:ZRR-results1}$\sim$~\ref{fig:ZRR-results4}, we show more qualitative results generated by PyNet~\cite{PyNet}, AWNet~\cite{AWNet}, MWISPNet~\cite{AIM2020ISP} and our models on the ZRR dataset. 

%

\begin{table*}[ht]
  \captionof{table}{Structure configuration of GCM model. GCM involves two components: SPN (left column) and GuideNet (right column). Except for the stride of the first convolutional layer in GuideNet is 2, the stride of other convolutional layers is 1.}%
  \label{tab:GCM_model}%
  \centering\noindent
  \centering%
  \vspace{-1mm}
  \begin{tabular}{cccc||cccc}%
    \specialrule{1pt}{0pt}{0pt}
    \multicolumn{4}{c||}{Spatially Preserving Network (SPN)} & \multicolumn{4}{c}{GuideNet} \\
    \hhline{----||----}
    Layer & Output size & Kernel size & Filter & Layer & Output size & Kernel size & Filter \\
    \hhline{====||====}
    Conv, ReLU                 & $448\times448$   & $1\times1$        & $5\rightarrow64$  
       & Conv                  & $222\times222$   & $7\times7$        & $8\rightarrow32$  \\
    {[Conv, ReLU]} $\times$3   & $448\times448$   & $1\times1$        & $64\rightarrow64$  
       & Conv, ReLU, Conv      & $222\times222$   & $3\times3$        & $32\rightarrow32$ \\
    Conv                       & $448\times448$   & $1\times1$        & $64\rightarrow3$   
       & Global Average Pooling  & $1\times1$     & - & - \\
                               &                       &                  &                    
       & Conv                  & $1\times1$       & $1\times1$        & $32\rightarrow64$ \\
    \specialrule{1pt}{0pt}{0pt}
  \end{tabular}
  \vspace{-4mm}
\end{table*}

\begin{table}[ht]
    \captionof{table}{Structure configuration of LiteISPNet. DWT and IWT denote discrete wavelet transform and inverse wavelet transform, respectively. RG denotes the residual group containing 4 residual channel attention blocks (RCABs).}
  \label{tab:LiteISPNet}%
  \centering\noindent
  \centering%
  \vspace{-2mm}
  \begin{tabular}{ccc}%
    \specialrule{1pt}{0pt}{0pt}
    \multicolumn{3}{c}{LiteISPNet} \\
    \hhline{---}
    Layer & Output size & Filter \\
    \hhline{===}
    Conv & $224\times224$ & $4\rightarrow64$     \\
    RG   & $224\times224$ & $64\rightarrow64$    \\
    DWT  & $112\times112$ & $64\rightarrow256$   \\ 
    \hhline{---}
    Conv & $112\times112$ & $256\rightarrow64$   \\
    RG   & $112\times112$ & $64\rightarrow64$    \\
    DWT  & $56\times56$   & $64\rightarrow256$    \\ 
    \hhline{---}
    Conv & $56\times56$   & $256\rightarrow128$   \\ 
    RG   & $56\times56$   & $128\rightarrow128$   \\ 
    DWT  & $28\times28$   & $128\rightarrow512$   \\ 
    \hhline{---}
    Conv & $28\times28$   & $512\rightarrow128$  \\
    RG   & $28\times28$   & $128\rightarrow128$  \\
    RG   & $28\times28$   & $128\rightarrow128$  \\
    Conv & $28\times28$   & $128\rightarrow512$  \\
     \hhline{---}
     IWT   & $56\times56$     & $512\rightarrow128$  \\
     RG    & $56\times56$     & $128\rightarrow128$ \\
     Conv  & $56\times56$     & $128\rightarrow256$ \\
     \hhline{---}
     IWT   & $112\times112$   & $256\rightarrow64$  \\
     RG    & $112\times112$   & $64\rightarrow64$  \\
     Conv  & $112\times112$   & $64\rightarrow256$  \\
     \hhline{---}
     IWT   & $224\times224$   & $256\rightarrow64$  \\
     RG    & $224\times224$   & $64\rightarrow64$  \\
     Conv  & $224\times224$   & $64\rightarrow64$  \\
     \hhline{---}
     Conv  & $224\times224$   & $64\rightarrow256$  \\
     PixelShuffle & $448\times448$   & $256\rightarrow64$  \\
     Conv  & $448\times448$   & $64\rightarrow3$  \\
    \specialrule{1pt}{0pt}{0pt}
  \end{tabular}
  \vspace{0mm}
\end{table}	    

\begin{table}[ht]
  \captionof{table}{Structure configuration of the discriminator. The kernel size of all convolutional layers is $4\times4$. The stride of the first three convolutional layers is 2, while the stride of the last two convolutional layers is 1.}
  \label{tab:Discriminator}%
  \centering\noindent
  \centering%
  \vspace{-1mm}
  \begin{tabular}{cccc}%
    \specialrule{1pt}{0pt}{0pt}
    \multicolumn{3}{c}{Discriminator} \\
    \hhline{---}
    Layer & Output size  & Filter \\
    \hhline{===}
    Conv, LeakyReLU & $224\times224$ & $3\rightarrow64$     \\
    Conv, BatchNorm, LeakyReLU & $112\times224$ & $64\rightarrow128$     \\
    Conv, BatchNorm, LeakyReLU & $56\times56$ & $128\rightarrow256$     \\
    Conv, BatchNorm, LeakyReLU & $55\times55$ & $256\rightarrow512$     \\
    Conv, BatchNorm, LeakyReLU & $54\times54$ & $512\rightarrow1$     \\
    \specialrule{1pt}{0pt}{0pt}
  \end{tabular}
  \vspace{-4mm}
\end{table}	

\begin{table}[ht] 
  \vspace{-2mm}
  \caption{Average results on all 24/100, 35/150 and 50/240 test pairs of SR-RAW dataset.
  Methods taking LR sRGB image as input are marked with $^\dagger$.
  The metrics are computed by \textit{Align GT with result}.}
  \label{tab:SRRAW}
  \centering\noindent
  \centering%
  \vspace{-4mm}
  \begin{center}
    \begin{tabular}{cc}
      \toprule
      Method & \tabincell{c}{ {PSNR}{$\uparrow$} {/} {SSIM}{$\uparrow$} {/} {LPIPS}{$\downarrow$}} \\
        \hline
        SRGAN$^\dagger$~\cite{SRResNet}        &  21.72 / 0.6917 / 0.394  \\
        ESRGAN$^\dagger$~\cite{ESRGAN}         &  21.85 / 0.6904 / 0.393 \\
        SPSR$^\dagger$~\cite{SPSR}             &  21.75 / 0.6692 / 0.427 \\
        RealSR$^\dagger$~\cite{RealSR}         &   21.89 / 0.6918 / 0.388 \\
        Zhang \etal~\cite{SRRAW}     & 21.97 / 0.7360 / 0.357   \\
        \hline
        Ours  & 22.50 / \textbf{0.7369} / 0.329 \\
        Ours (GAN)    & \textbf{22.56} / 0.7341 / \textbf{0.323}  \\
      \bottomrule
    \end{tabular}
  \end{center}
  \vspace{-7mm}
\end{table}

\begin{table}[ht] 
  \vspace{-2mm}
  \caption{Quantitative results for re-splitting the train/test set of ZRR dataset.
  The metrics are computed by \textit{Align GT with result}.}
  \label{tab:ZRR}
  \centering\noindent
  \centering%
  \vspace{-4mm}
  \begin{center}
    \begin{tabular}{cc}
      \toprule
      Method & \tabincell{c}{ {PSNR}{$\uparrow$} {/} {SSIM}{$\uparrow$} {/} {LPIPS}{$\downarrow$}} \\
        \hline
        PyNet~\cite{PyNet}     & 22.67 / 0.8535 / 0.149   \\
        AWNet (raw)~\cite{AWNet}     & 22.83 / 0.8513 / 0.160        \\
        AWNet (demosaicked)~\cite{AWNet}     & 22.68 / 0.8447 / 0.173        \\
        MWISPNet~\cite{AIM2020ISP}     & 23.00 / 0.8530 / 0.166        \\
        \hline 
        Ours (LiteISPNet)   & \textbf{23.31} / \textbf{0.8747} / \textbf{0.131} \\
      \bottomrule
    \end{tabular}
  \end{center}
  \vspace{-4mm}
\end{table}

\clearpage

\begin{figure*}[t] 
 \centering
 \subfigure[Raw image] 
{
	\begin{minipage}{.15\linewidth}
		\centering
		\includegraphics[width=\linewidth]{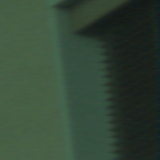}

		\includegraphics[width=\linewidth]{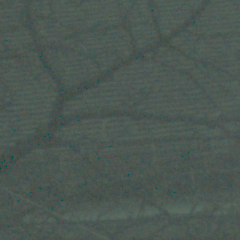}
        
		\includegraphics[width=\linewidth]{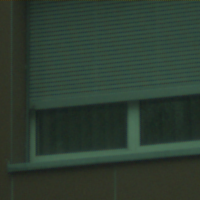}
     \vspace{-0.2cm}
	\end{minipage}
}%
 \subfigure[SIFT (baseline)]
 {
   \begin{minipage}{.15\linewidth}
     \centering
     \includegraphics[width=\linewidth]{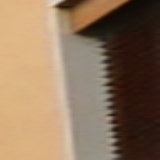}

     \includegraphics[width=\linewidth]{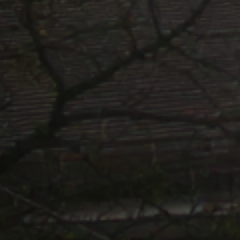}
    
     \includegraphics[width=\linewidth]{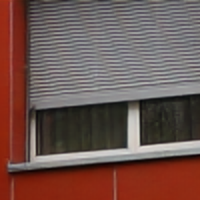}
     \vspace{-0.2cm}
   \end{minipage}
 }%
 \subfigure[Align $\mathbf{y}$ to $\hat{\mathbf{y}}$]
 {
   \begin{minipage}{.15\linewidth}
     \centering
     \includegraphics[width=\linewidth]{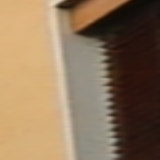}

     \includegraphics[width=\linewidth]{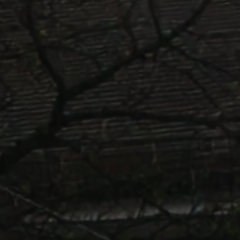}

     \includegraphics[width=\linewidth]{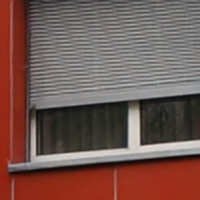}
     \vspace{-0.2cm}
   \end{minipage}
 }%
 \subfigure[Align $\mathbf{y}$ to $\hat{\mathbf{x}}$]
 {
   \begin{minipage}{.15\linewidth}
     \centering
     \includegraphics[width=\linewidth]{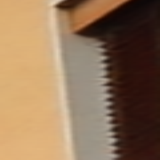}

     \includegraphics[width=\linewidth]{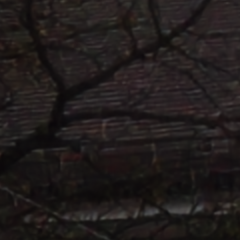}

     \includegraphics[width=\linewidth]{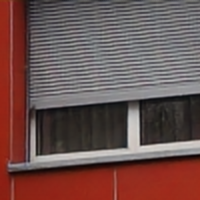}
     \vspace{-0.2cm}
   \end{minipage}
 }%
 \subfigure[Align $\mathbf{y}$ to $\tilde{\mathbf{y}}$ (Ours)]
 {
   \begin{minipage}{.15\linewidth}
     \centering
     \includegraphics[width=\linewidth]{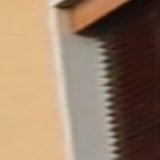}

     \includegraphics[width=\linewidth]{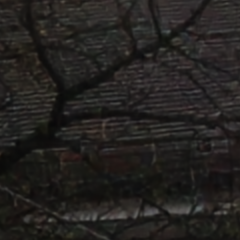}

     \includegraphics[width=\linewidth]{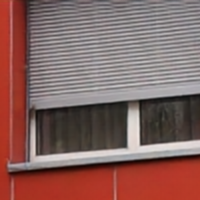}
     \vspace{-0.2cm}
   \end{minipage}
 }%
 \subfigure[GT]
 {
   \begin{minipage}{.15\linewidth}
     \centering
     \includegraphics[width=\linewidth]{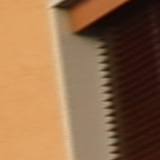}

     \includegraphics[width=\linewidth]{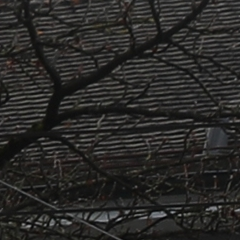}

     \includegraphics[width=\linewidth]{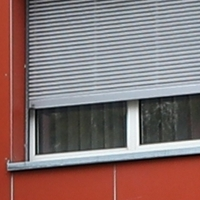}
     \vspace{-0.2cm}
   \end{minipage}
 }%

 \caption{Visual results of LiteISPNet output $\hat{\mathbf{y}}$. (b)$\sim$(e) denote different alignment strategies. Our method (e) performs favorably against other alignment strategies.}
 \label{fig:ZRR-ablation-alignment}
 \end{figure*}    
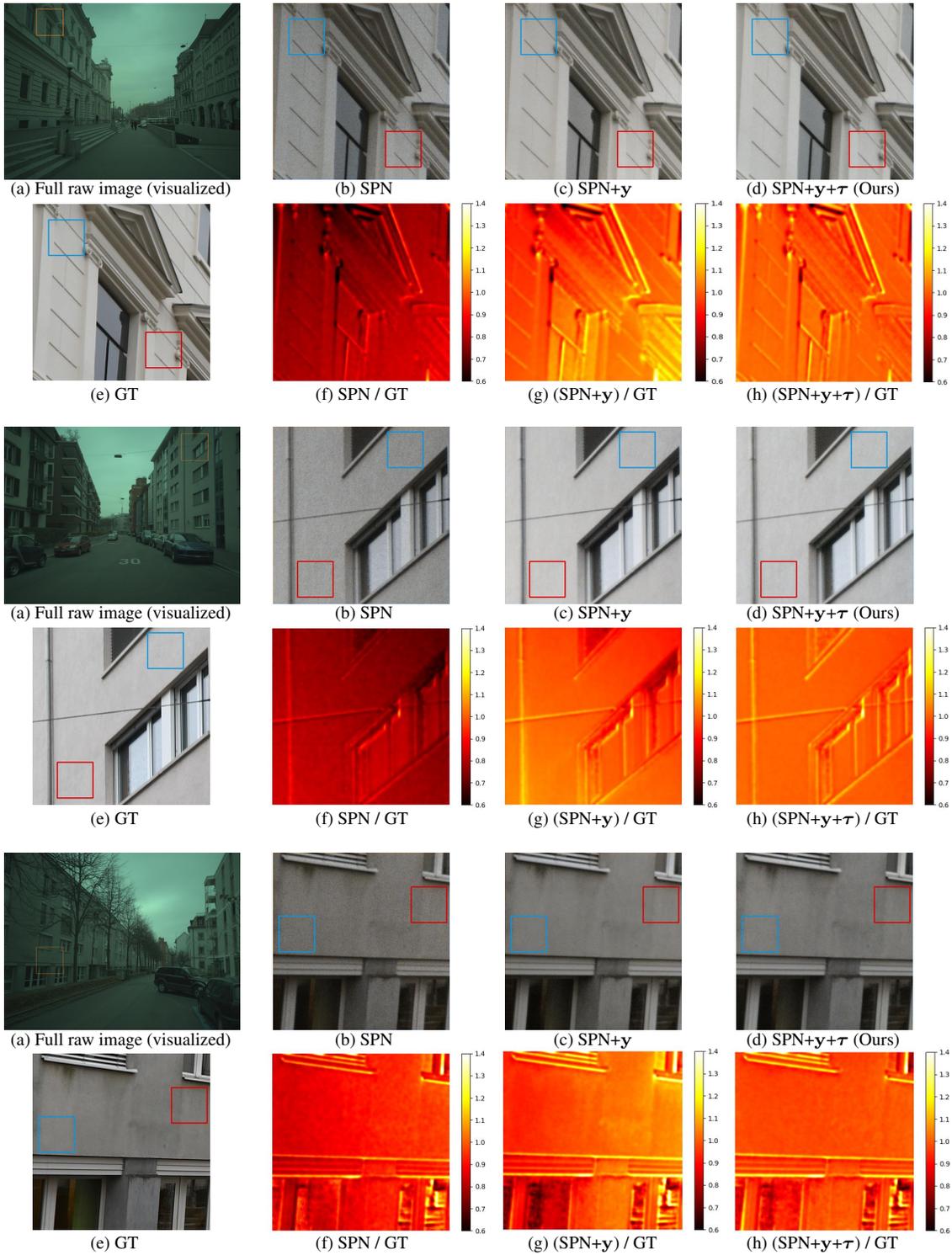
\begin{figure*}[h]
	\centering
	\vspace{-2mm}
  \begin{overpic}[width=.9\linewidth]{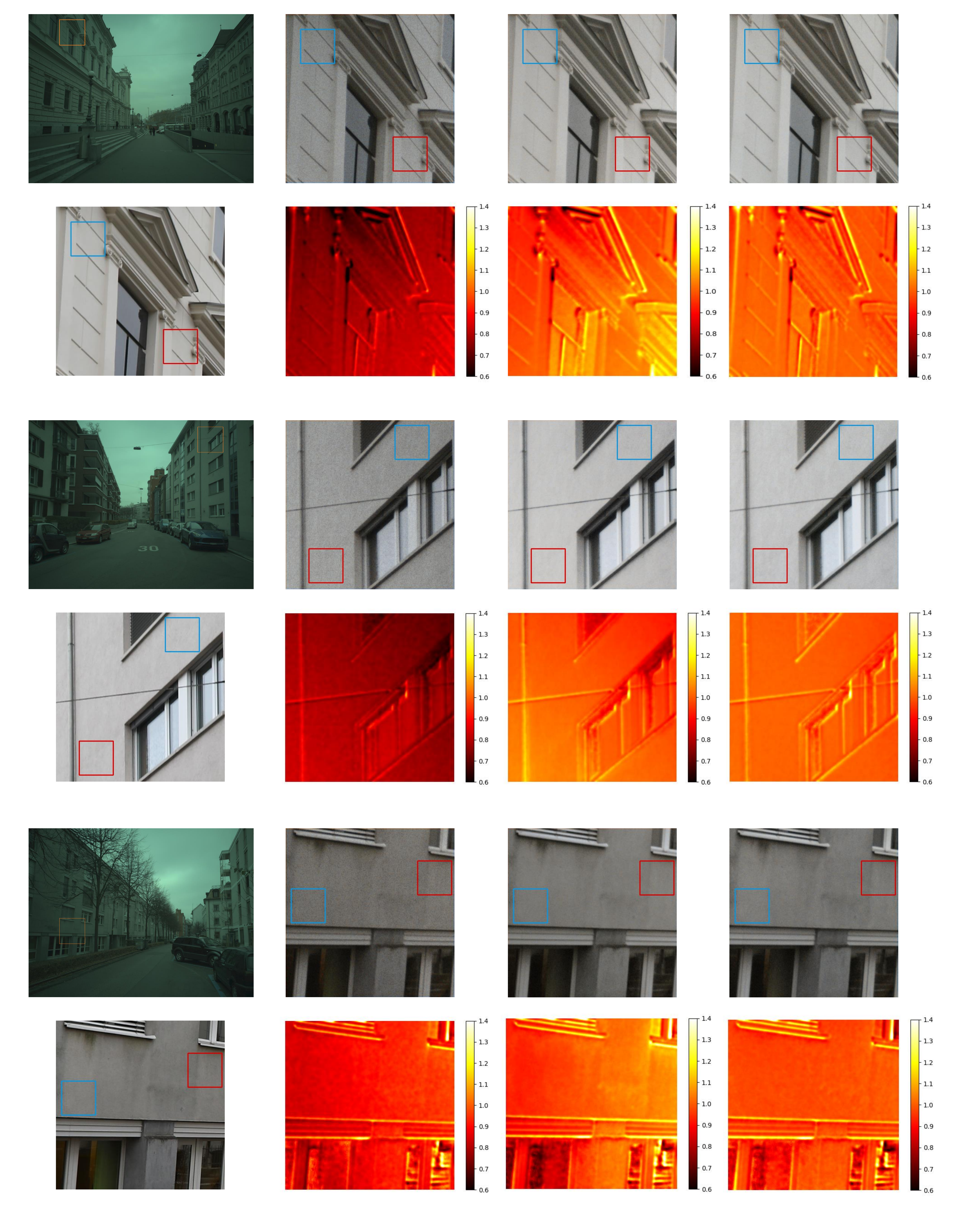}
		\put(2.8, 84){\footnotesize (a) Full raw image (visualized)}
    \put(28, 84){\footnotesize (b) SPN}
    \put(45, 84){\footnotesize (c) SPN+$\mathbf{y}$}
    \put(60, 84){\footnotesize (d) SPN+$\mathbf{y}$+$\bm{\tau}$ (Ours)}
    \put(9, 68){\footnotesize (e) GT}
    \put(26.5, 68){\footnotesize (f) SPN / GT}
    \put(43, 68){\footnotesize (g) (SPN+$\mathbf{y}$) / GT}
    \put(60, 68){\footnotesize (h) (SPN+$\mathbf{y}$+$\bm{\tau}$) / GT}
    \put(2.8, 51){\footnotesize (a) Full raw image (visualized)}
    \put(28, 51){\footnotesize (b) SPN}
    \put(45, 51){\footnotesize (c) SPN+$\mathbf{y}$}
    \put(60, 51){\footnotesize (d) SPN+$\mathbf{y}$+$\bm{\tau}$ (Ours)}
    \put(9, 35){\footnotesize (e) GT}
    \put(26.5, 35){\footnotesize (f) SPN / GT}
    \put(43, 35){\footnotesize (g) (SPN+$\mathbf{y}$) / GT}
    \put(60, 35){\footnotesize (h) (SPN+$\mathbf{y}$+$\bm{\tau}$) / GT}
    \put(2.8, 17.8){\footnotesize (a) Full raw image (visualized)}
    \put(28, 17.8){\footnotesize (b) SPN}
    \put(45, 17.8){\footnotesize (c) SPN+$\mathbf{y}$}
    \put(60, 17.8){\footnotesize (d) SPN+$\mathbf{y}$+$\bm{\tau}$ (Ours)}
    \put(9, 2){\footnotesize (e) GT}
    \put(26.5, 2){\footnotesize (f) SPN / GT}
    \put(43, 2){\footnotesize (g) (SPN+$\mathbf{y}$) / GT}
    \put(60, 2){\footnotesize (h) (SPN+$\mathbf{y}$+$\bm{\tau}$) / GT}
	\end{overpic}
	\vspace{-3mm}
	\caption{Visual results of GCM output $\tilde{\mathbf{y}}$. (b)$\sim$(d) denote the results using different GCM components. (f)$\sim$(h) denote the illumination ratio between (b)$\sim$(d) and GT, respectively. With the guidence of $\mathbf{y}$, the color of GCM output $\tilde{\mathbf{y}}$ in (c) and (d) is closer to the target sRGB image. Dark corner can be observed in (b) and (c). In (b) and (c), the patch in the blue box is darker than the patch in the red box. But in (d) and (e), the patch in different boxes has similar illumination. The phenomenon can be seen more clearly in (f)$\sim$(h).}
	\label{fig:ZRR-ablation-GCM}
	\vspace{-0.5cm}
\end{figure*}
     \begin{figure*}[t] 
     \centering
     \subfigure[Bicubic$^\dagger$]
     {
      \begin{minipage}{.19\linewidth}
         \centering
         \includegraphics[width=\linewidth]{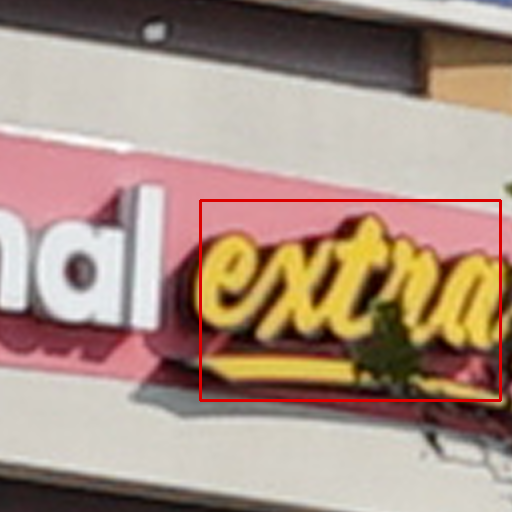}
         \vspace{-0.3cm}
      \end{minipage}
     }%
     \subfigure[SRGAN$^\dagger$~\cite{SRResNet}]
     {
      \begin{minipage}{.19\linewidth}
         \centering
         \includegraphics[width=\linewidth]{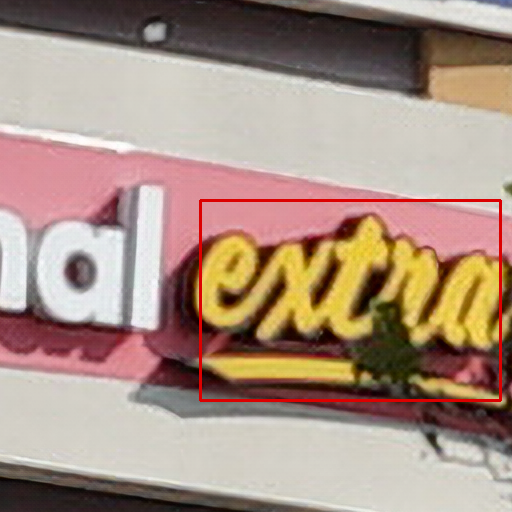}
         \vspace{-0.3cm}
      \end{minipage}
     }%
     \subfigure[ESRGAN$^\dagger$~\cite{ESRGAN}]
     {
      \begin{minipage}{.19\linewidth}
         \centering
         \includegraphics[width=\linewidth]{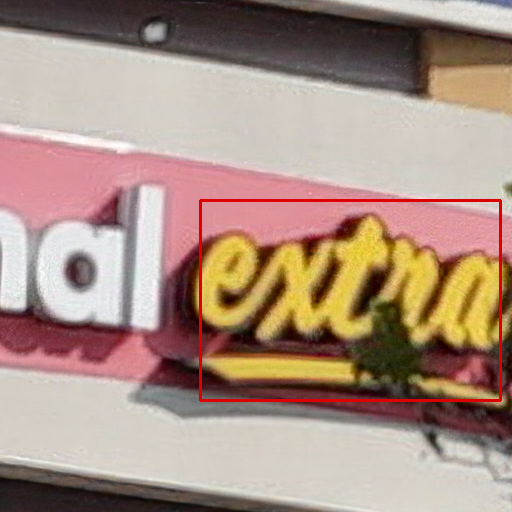}
         \vspace{-0.3cm}
      \end{minipage}
     }%
     \subfigure[SPSR$^\dagger$~\cite{SPSR}]
     {
      \begin{minipage}{.19\linewidth}
         \centering
         \includegraphics[width=\linewidth]{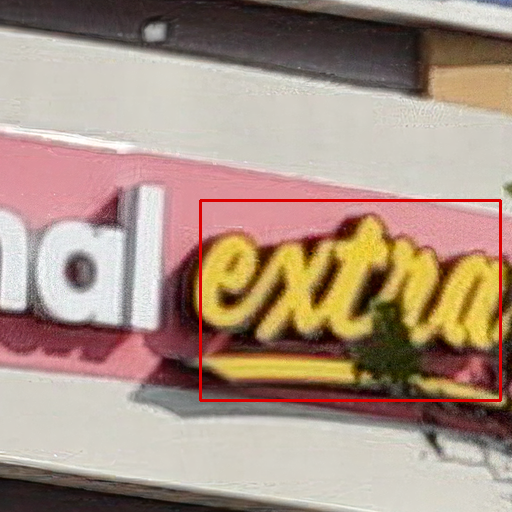}
         \vspace{-0.3cm}
      \end{minipage}
     }%
     \subfigure[RealSR$^\dagger$~\cite{RealSR}]
     {
      \begin{minipage}{.19\linewidth}
         \centering
         \includegraphics[width=\linewidth]{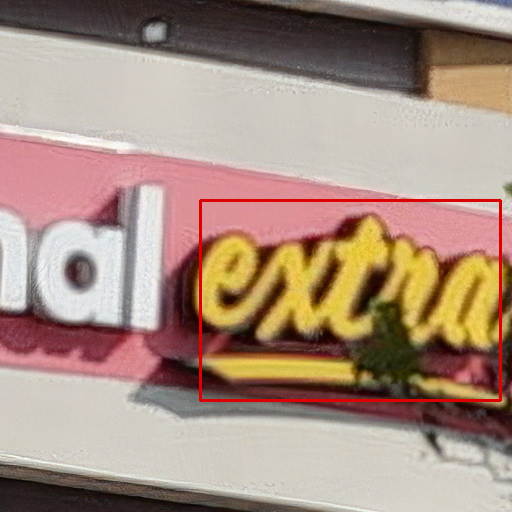}
         \vspace{-0.3cm}
      \end{minipage}
     }%

     \subfigure[Raw image (visualized)]
     {
      \begin{minipage}{.19\linewidth}
         \centering
         \includegraphics[width=\linewidth]{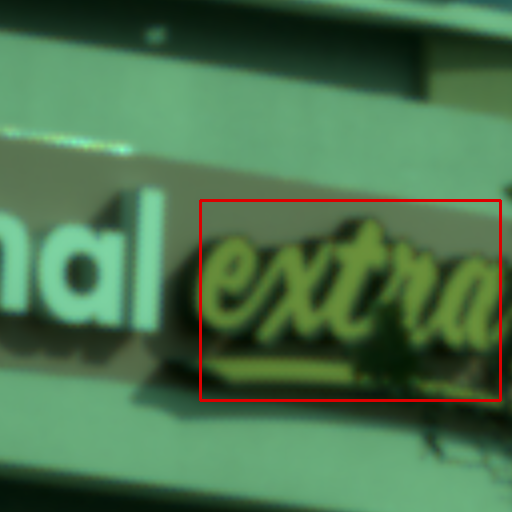}
         \vspace{-0.3cm}
      \end{minipage}
     }%
     \subfigure[Zhang \etal~\cite{SRRAW}]
     {
      \begin{minipage}{.19\linewidth}
         \centering
         \includegraphics[width=\linewidth]{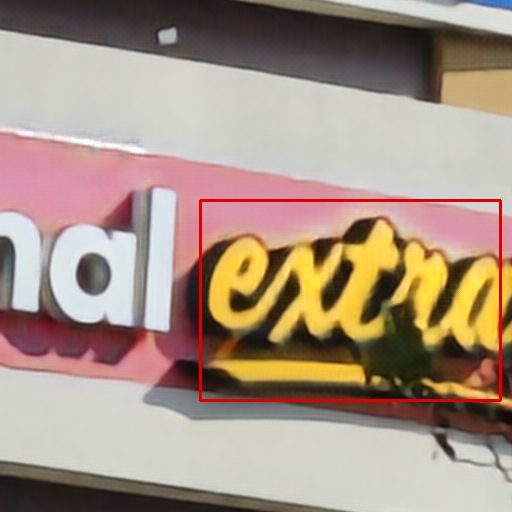}
         \vspace{-0.3cm}
      \end{minipage}
     }%
     \subfigure[Ours]
     {
      \begin{minipage}{.19\linewidth}
         \centering
         \includegraphics[width=\linewidth]{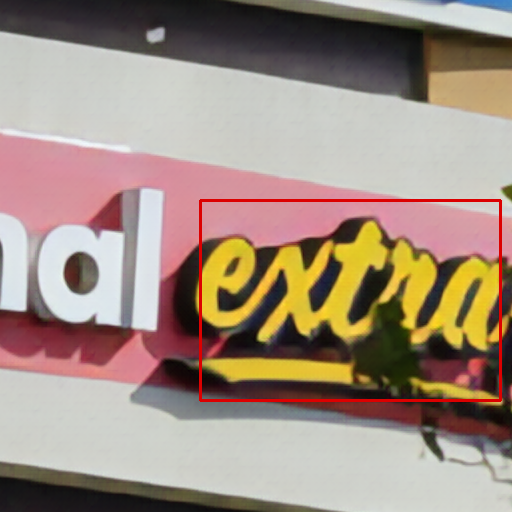}
         \vspace{-0.3cm}
      \end{minipage}
     }%
     \subfigure[Ours (GAN)]
     {
      \begin{minipage}{.19\linewidth}
         \centering
         \includegraphics[width=\linewidth]{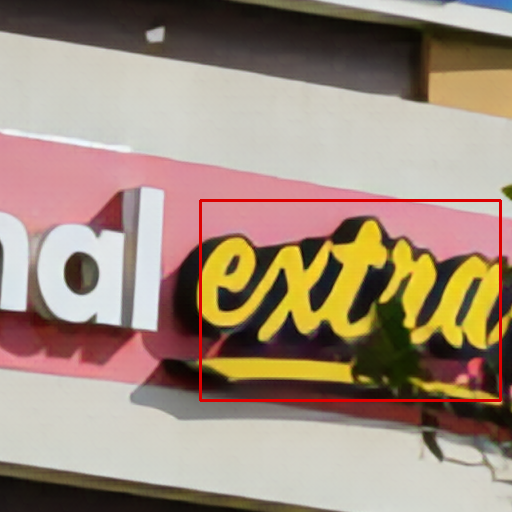}
         \vspace{-0.3cm}
      \end{minipage}
     }%
     \subfigure[GT] 
     {
      \begin{minipage}{.19\linewidth}
         \centering
         \includegraphics[width=\linewidth]{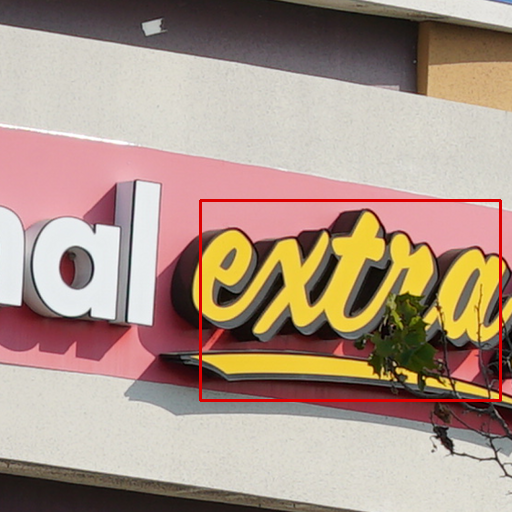}
         \vspace{-0.3cm}
      \end{minipage}
     }%
     \caption{Visual comparison on SR-RAW dataset. $^\dagger$ means that the result is obtained given LR sRGB image as input.
     The edges of our results are sharper. It can be clearly observed in the red box.}
     \label{fig:SRRAW-result5}
  \end{figure*}

    %
    \begin{figure*}[t] 
    \centering
    \subfigure[Raw image (visualized)]
    {
      \begin{minipage}{.25\linewidth}
        \centering
        \includegraphics[width=\linewidth]{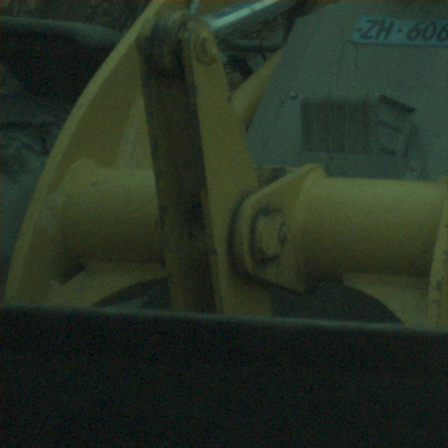}
        \vspace{-0.3cm}
      \end{minipage}
    }%
    \subfigure[PyNet~\cite{PyNet}]
    {
      \begin{minipage}{.25\linewidth}
        \centering
        \includegraphics[width=\linewidth]{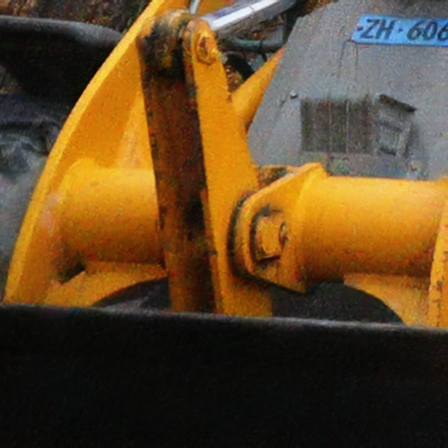}
        \vspace{-0.3cm}
      \end{minipage}
    }%
    \subfigure[AWNet (raw)~\cite{AWNet}]
    {
      \begin{minipage}{.25\linewidth}
        \centering
        \includegraphics[width=\linewidth]{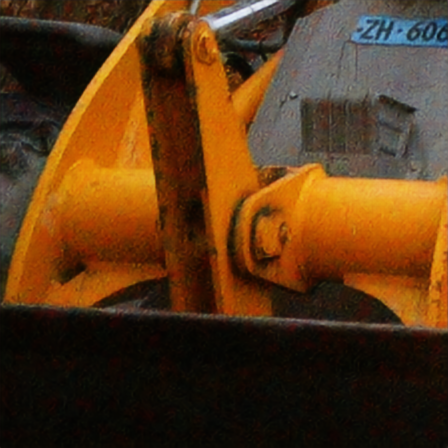}
        \vspace{-0.3cm}
      \end{minipage}
    }%
    \subfigure[AWNet (demosaicked)~\cite{AWNet}]
    {
      \begin{minipage}{.25\linewidth}
        \centering
        \includegraphics[width=\linewidth]{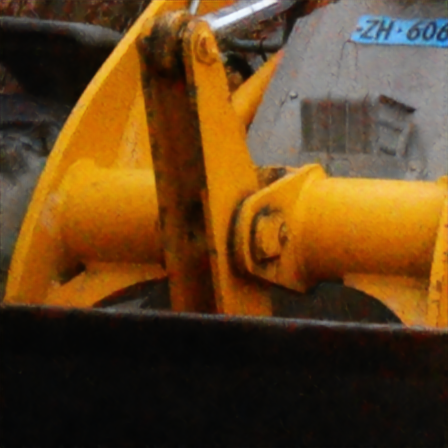}
        \vspace{-0.3cm}
      \end{minipage}
    }%
    
    \subfigure[MW-ISPNet (GAN)~\cite{AIM2020ISP}]
    {
      \begin{minipage}{.25\linewidth}
        \centering
        \includegraphics[width=\linewidth]{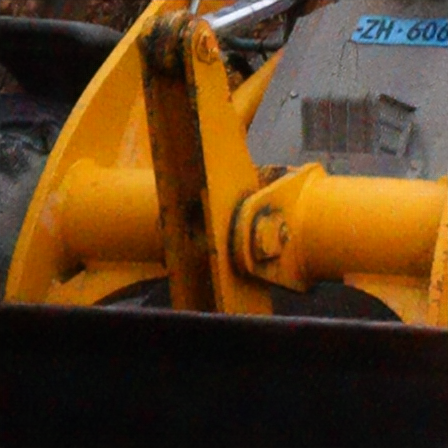}
        \vspace{-0.3cm}
      \end{minipage}
    }%
    \subfigure[Ours (LiteISPNet)]
    {
      \begin{minipage}{.25\linewidth}
        \centering
        \includegraphics[width=\linewidth]{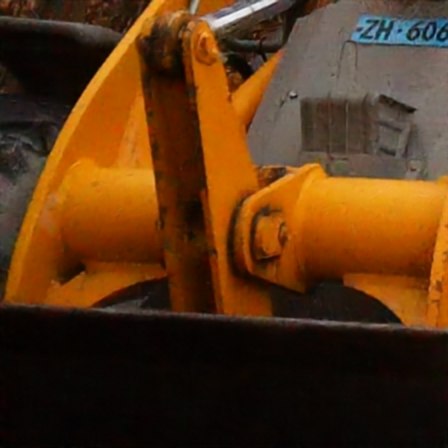}
        \vspace{-0.3cm}
      \end{minipage}
    }%
    \subfigure[Ours (LiteISPGAN)]
    {
      \begin{minipage}{.25\linewidth}
        \centering
        \includegraphics[width=\linewidth]{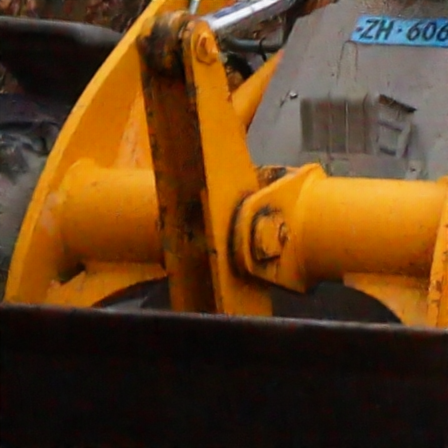}
        \vspace{-0.3cm}
      \end{minipage}
    }%
    \subfigure[GT]
    {
      \begin{minipage}{.25\linewidth}
        \centering
        \includegraphics[width=\linewidth]{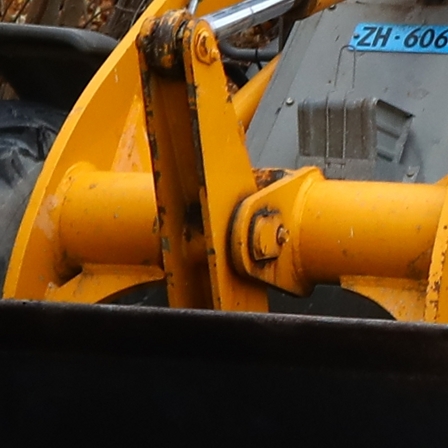}
        \vspace{-0.3cm}
      \end{minipage}
    }%
    \caption{Visual comparisons on ZRR dataset. Our results have less noise.}
    \label{fig:ZRR-results1}
    \end{figure*}    
    \begin{figure*}[t] 
    \centering
    \subfigure[Raw image (visualized)] 
    {
      \begin{minipage}{.25\linewidth}
        \centering
        \includegraphics[width=\linewidth]{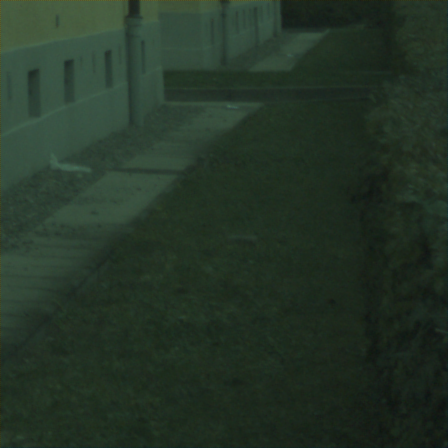}
        \vspace{-0.3cm}
      \end{minipage}
    }%
    \subfigure[PyNet~\cite{PyNet}]
    {
      \begin{minipage}{.25\linewidth}
        \centering
        \includegraphics[width=\linewidth]{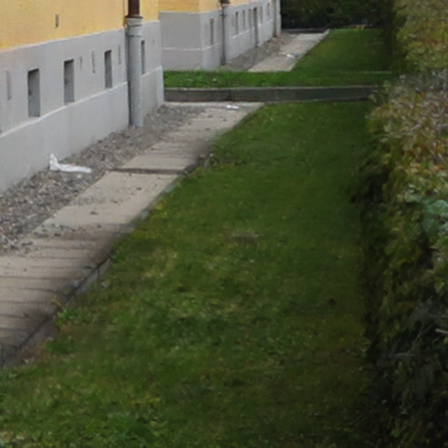}
        \vspace{-0.3cm}
      \end{minipage}
    }%
    \subfigure[AWNet (raw)~\cite{AWNet}]
    {
      \begin{minipage}{.25\linewidth}
        \centering
        \includegraphics[width=\linewidth]{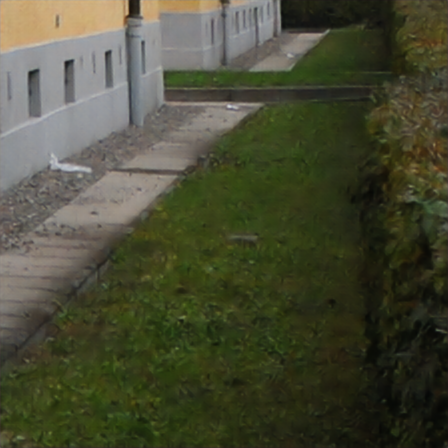}
        \vspace{-0.3cm}
      \end{minipage}
    }%
    \subfigure[AWNet (demosaicked)~\cite{AWNet}]
    {
      \begin{minipage}{.25\linewidth}
        \centering
        \includegraphics[width=\linewidth]{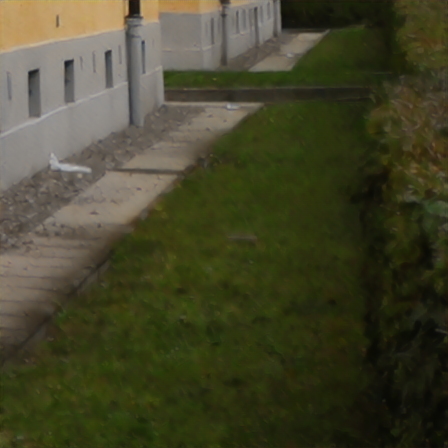}
        \vspace{-0.3cm}
      \end{minipage}
    }%
    
    \subfigure[MW-ISPNet (GAN)~\cite{AIM2020ISP}]
    {
      \begin{minipage}{.25\linewidth}
        \centering
        \includegraphics[width=\linewidth]{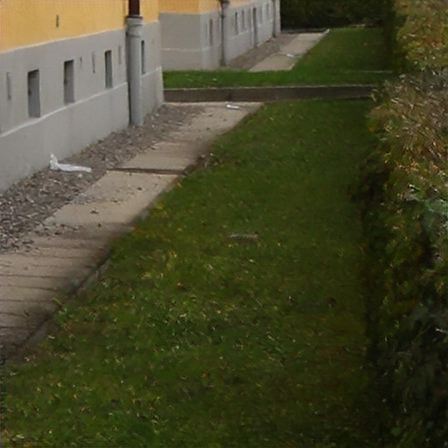}
        \vspace{-0.3cm}
      \end{minipage}
    }%
    \subfigure[Ours (LiteISPNet)]
    {
      \begin{minipage}{.25\linewidth}
        \centering
        \includegraphics[width=\linewidth]{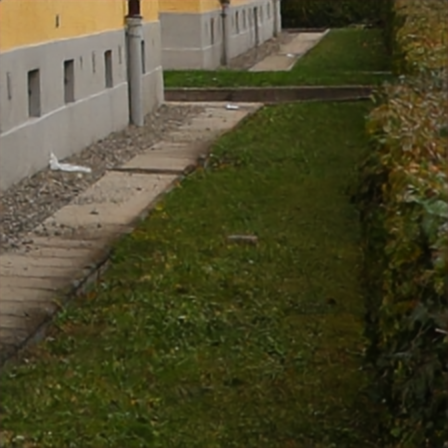}
        \vspace{-0.3cm}
      \end{minipage}
    }%
    \subfigure[Ours (LiteISPGAN)]
    {
      \begin{minipage}{.25\linewidth}
        \centering
        \includegraphics[width=\linewidth]{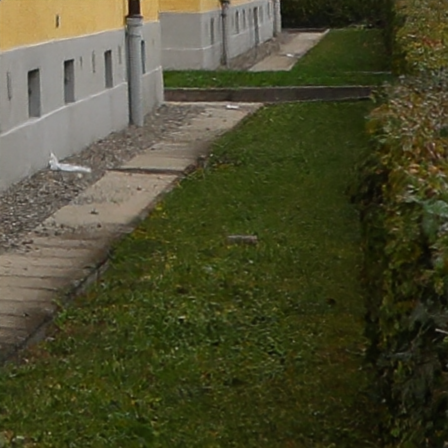}
        \vspace{-0.3cm}
      \end{minipage}
    }%
    \subfigure[GT]
    {
      \begin{minipage}{.25\linewidth}
        \centering
        \includegraphics[width=\linewidth]{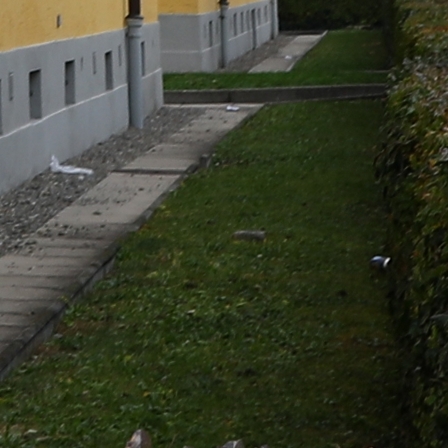}
        \vspace{-0.3cm}
      \end{minipage}
    }%
    \caption{Visual comparisons on ZRR dataset. 
    Our results have richer textures on the grass.}
    \label{fig:ZRR-results2}
    \end{figure*}  
%
%
%
%
%

\end{document}